%% file: main.tex
\documentclass[10pt,twocolumn,letterpaper]{article}

\usepackage[cvprfinal]{cvpr}      

\usepackage{colortbl}
\usepackage{graphicx}
\usepackage{amsmath}
\usepackage{amssymb}
\usepackage{booktabs}
\usepackage{algorithm}
\usepackage{algorithmic}
\usepackage{subcaption}
\usepackage{bbding}
\usepackage{xcolor}
\usepackage{dsfont}
\usepackage{bm}
\usepackage[pagebackref,breaklinks,colorlinks]{hyperref}
\usepackage[capitalize]{cleveref}
\definecolor{cyan}{cmyk}{.3,0,0,0}
\usepackage[capitalize]{cleveref}
\crefname{section}{Sec.}{Secs.}
\Crefname{section}{Section}{Sections}
\Crefname{table}{Table}{Tables}
\crefname{table}{Tab.}{Tabs.}

\newcommand{\modelname}{Uncertainty-aware Pseudo-label-filtering Adaptation}
\newcommand{\shortname}{UPA}
\newcommand{\selectionname}{Adaptive Pseudo-label Selection}
\newcommand{\selectshortname}{APS}
\newcommand{\clname}{Class-Aware Contrastive Learning}
\newcommand{\clshortname}{CACL}
\newcommand{\xchen}[1]{\textcolor{black}{#1}}

\newcommand{\mytext}[1]{\noindent\textbf{#1}}

\begin{document}

\title{Uncertainty-Aware Pseudo-Label Filtering for Source-Free Unsupervised Domain Adaptation}

\author{Xi Chen\\
Harbin Institute of Technology\\
{\tt\small xichen98cn@gmail.com}
\and
Haosen Yang\\
University of Surrey\\
{\tt\small haosen.yang.6@gmail.com}
\and
Huicong Zhang\\
Harbin Institute of Technology\\
{\tt\small huicongzhang@stu.hit.edu.cn}
\and
Hongxun Yao$^\dag$\\
Harbin Institute of Technology\\
{\tt\small h.yao@hit.edu.cn}
\and
Xiatian Zhu\\
University of Surrey\\
{\tt\small eddy.zhuxt@gmail.com}
}
\maketitle

{\let\thefootnote\relax\footnotetext{ ~$^\dag$ Corresponding authors: 
\href{mailto:h.yao@hit.edu.cn}{\color{black}{h.yao@hit.edu.cn}}}}

\begin{abstract}
Source-free unsupervised domain adaptation (SFUDA) aims to enable the utilization of a pre-trained source model in an unlabeled target domain without access to source data. Self-training is a way to solve SFUDA, where confident target samples are iteratively selected as pseudo-labeled samples to guide target model learning. However, prior heuristic noisy pseudo-label filtering methods all involve introducing extra models, which are sensitive to model assumptions and may introduce additional errors or mislabeling. 
In this work, we propose a method called \modelname{} (\shortname{}) to efficiently address this issue in a coarse-to-fine manner. Specially, we first introduce a sample selection module named \selectionname{} (\selectshortname), which is responsible for filtering noisy pseudo labels. The \selectshortname{} utilizes a simple sample uncertainty estimation method by aggregating knowledge from neighboring samples and confident samples are selected as clean pseudo-labeled. Additionally, we incorporate \clname{} (\clshortname{}) to mitigate the memorization of pseudo-label noise by learning robust pair-wise representation supervised by pseudo labels. Through extensive experiments conducted on three widely used benchmarks, we demonstrate that our proposed method achieves competitive performance on par with state-of-the-art SFUDA methods. Code is available at \url{https://github.com/chenxi52/UPA}.
\end{abstract}

\begin{figure*}[ht]
\centering
\begin{subfigure}[b]{0.45\textwidth}{
\includegraphics[width=\textwidth]{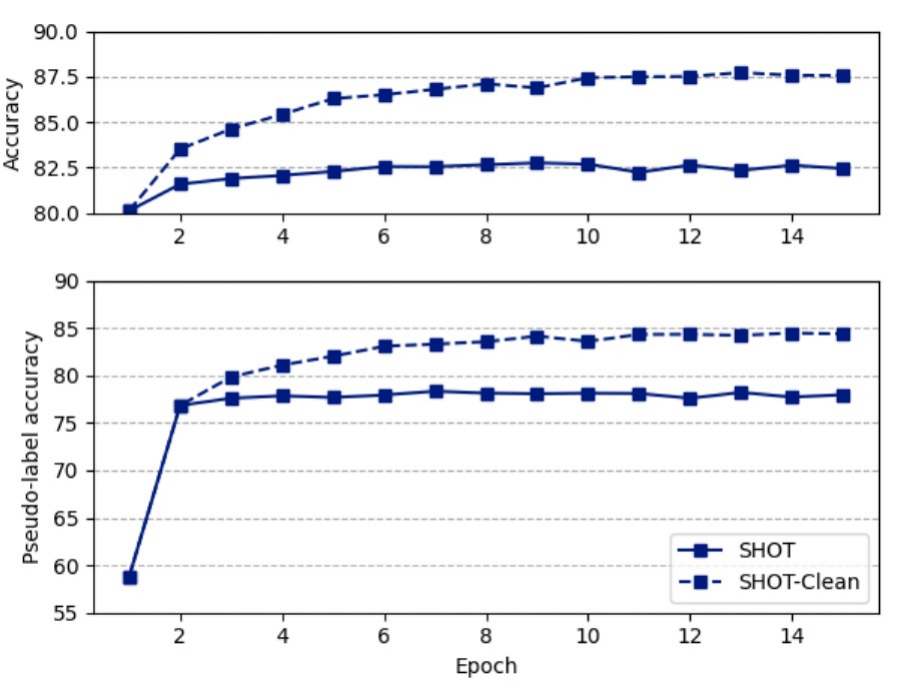}
\caption{}
\label{fig:fig1}
}
\end{subfigure}
\hfill
\begin{subfigure}[b]{0.45\textwidth}
\centering
    \raisebox{0.8cm}{
        \includegraphics[width=\textwidth]{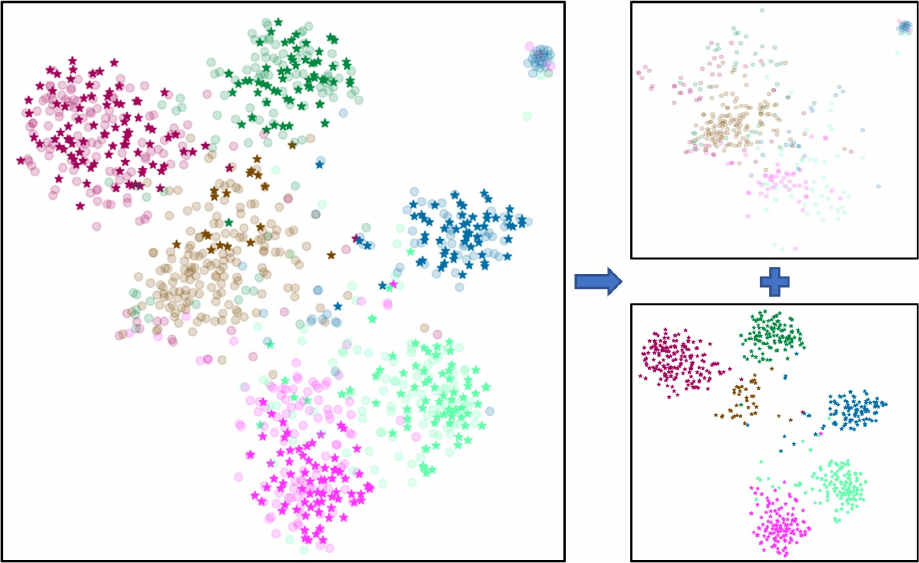}
        }
    \caption{}
    \label{fig:fig1_2}
\end{subfigure}
\caption{
    (a). Evolution of adaptation accuracy (top) and pseudo-label accuracy (bottom) for the SHOT\cite{SHOT} and  SHOT-clean (ideal simulation supervised by manually selected correct pseudo labels) methods on the VisDA-C~\cite{Visda-c} dataset.
    (b). 
    Feature visualization from 6 interval-selected classes of VisDA-C using t-SNE~\cite{t-sne}, 
    showcasing the differentiation between correct pseudo labels (represented by opaque stars) and incorrect pseudo labels (represented by semi-transparent dots) in the SFUDA setting.
    }
\end{figure*}

\section{Introduction}

The increasingly urgent needs for information security and privacy prohibit the sharing of data across different domains in practice, posing challenges for knowledge transfer. Consequently, model transfer has emerged as a promising approach, attracting recent research attention in the field of source-free unsupervised domain adaptation (SFUDA) tasks.
The goal of SFUDA is to transfer a pre-trained model, originally trained on a labeled training set from a source domain, to a target domain that only provides access to unlabeled training data. In this scenario, the source domain training data is unavailable, reflecting a more realistic setting.

Compared to the task of unsupervised domain adaptation (UDA), the key distinction in the SFUDA task lies in the absence of source data. This fundamental difference renders standard UDA methods~\cite{D-CORAL, MDD, MADA, CAN} inapplicable to the SFUDA setting, as they rely on data from both the source and target domains.
To address this challenge, SFUDA methods~\cite{3C-GAN, CPGA, CDCL, SHOT, U-SFAN, CoWA-JMDS} leverage source model and target data information mining. Typically, SFUDA methods aim to maximize the utilization of class-aware information, given that only target data is accessible. Consequently, 
self-training SFUDA methods involve generating pseudo-class prototypes~\cite{CPGA, 3C-GAN, CDCL} and assigning pseudo labels~\cite{SHOT, CoWA-JMDS} based on these prototypes. However, the pseudo-labeling method is susceptible to noisy predictions, which can undermine their accuracy and reliability.
As illustrated in Fig.~\ref{fig:fig1}, there is a strong correlation between adaptation accuracy and the correctness rate of pseudo labels. Notably, SHOT-clean achieves improved adaptation accuracy by manually selecting correct pseudo labels. Consequently, significant efforts~\cite{VDM-DA, D-MCD, SSNLL} have been devoted to mitigating the impact of noisy pseudo labels.
Despite their strong performance, these methods are suboptimal in measuring the quality of pseudo labels since they rely on additional models, which introduce additional modeling errors due to sensitive model assumptions.

Based on the presence of similar intra-class features in both the source and target domains, we visualize the target features generated by the source model in Fig.~\ref{fig:fig1_2}. Notably, the clean pseudo-labeled features, represented by stars, are clearly clustered, indicating the presence of domain-shared information. 
We posit that target features exhibiting such distinct clustering patterns in the feature space are more resilient to domain shifts. 
Motivated by this hypothesis, 
we propose a novel pseudo-label denoising method called \modelname{} (\shortname{}). 
We estimate sample uncertainty by evaluating the degree of clustering and introduce the \selectionname{} (\selectshortname) method, which iteratively assesses the uncertainty of a target sample by considering the pseudo labels of its neighboring samples in the spatial domain. This facilitates subsequent sample selection and enables the identification of clean pseudo labels. Importantly, the \selectshortname{} not only denoises pseudo labels but is also model-free, avoiding additional model bias.
To further reduce pseudo-label noise in a more refined manner, we introduce \clshortname{}. This method ensures intra-class semantic consistency in the target data by applying category-wise contrastive learning~\cite{SupCon} to clean pseudo-labeled samples. Positive pairs consist of samples belonging to the same class, while negative pairs are composed of samples from different classes.

In a nutshell, our approach involves a source-model-centric uncertainty estimation and pseudo-label-guided class-wise contrastive learning. 
The target data is partitioned into two sets (clean pseudo-labeled vs. noisy pseudo-labeled).
This partitioning helps guide the adaptation process by utilizing clean pseudo-label supervision and further enhances the robustness of model representation. 
Our method offers the advantages of being: (1) model-free for pseudo-label denoising, (2) easy-to-use (i.e. self-learning way), and (3) coarse to fine (i.e.from pseudo-label denoising to preventing noisy memorization ).

Our {\bf contributions} can be summarized as follows:

\begin{itemize}
\item We propose a pseudo-label filtering framework called \shortname{} for the SFUDA task, which effectively addresses the challenge of high pseudo-label noise without requiring additional models for uncertainty estimation. We introduce APS, an uncertainty-aware mechanism that efficiently filters out noisy pseudo labels. It iteratively separates target samples into pseudo-labeled and unlabeled sets.

\item 
We introduce a semantic consistency learning strategy that incorporates class-wise contrastive learning for the pseudo-labeled set. This strategy prevents the memorization of label noise and ensures robust learning.

\item We conduct extensive experiments on three standard datasets (Office, Office-Home, and VisDA-C) to demonstrate the performance advantage of our methods over existing alternatives.

\end{itemize}

\section{Related Work}
\subsection{Unsupervised Domain Adaptation (UDA)}
UDA methods have been developed to reduce the distribution discrepancy between the source and target domains. These methods can broadly be categorized into two primary groups.

The first group of methods~\cite{MMD, MDD, FIxBi, CORAL, D-CORAL, CDAN, CAN, BCDM, MUDA, MADAN, DANN, ASOS, GAM} focuses on minimizing various metrics of cross-domain distribution divergence by performing feature transformation. 
MMD~\cite{MMD} minimizes the Maximum Mean Discrepancy between domains while MDD~\cite{MDD} targets the reduction of Margin Disparity Discrepancy. Meanwhile, D-CORAL~\cite{D-CORAL} and CORAL~\cite{CORAL} mitigate domain discrepancies by correlation alignment.~\cite{DANN, ASOS, GAM} learns the domain-invariant features by inversely back-propagating the gradients from the loss of the domain classifier.
Their objective is to reduce the domain gap and enhance adaptation performance.
The second group of methods~\cite{SRDC, RSDA, SimNet, SACR} adopts a self-learning scheme. In this scheme, labeled data from the source domain is utilized to label the unlabeled data, followed by self-training or pseudo-labeling methods. 
SRDC\cite{SRDC} employs discriminative clustering on the target data, revealing the inherent target discrimination. RSDA~\cite{RSDA} introduces a robust pseudo-label loss within the spherical feature space. Meanwhile, SimNet~\cite{SimNet} classifies by measuring the similarity between prototype representations for each category.

However, it is important to note that these existing UDA methods assume the availability of labeled source domain training data, which is not the case in the SFUDA problem addressed in this study.

\subsection{Source-Free Unsupervised Domain Adaptation (SFUDA)}
SFUDA focuses on transferring knowledge from a pre-trained source model to a target domain without the need for labeled source data. Existing SFUDA methods can be categorized into two main groups.

The first group of methods~\cite{3C-GAN, CPGA, CDCL, A2Net} focuses on "cross-domain alignment" strategies. These methods reproduce fake source knowledge under source domain hypotheses or by utilizing source model information. For example, 3C-GAN~\cite{3C-GAN} generates source-style images during adaptation, CPGA~\cite{CPGA} creates avatar feature prototypes using the source classifier, and CDCL~\cite{CDCL} utilizes weights of the source classifier as source anchors. A2Net~\cite{A2Net} employs a dual-classifier design, with one of the classifiers being the frozen source classifier, to achieve adversarial domain-level alignment.

The second group of methods~\cite{SHOT, D-MCD, VDM-DA, SSNLL, NRC, CoWA-JMDS, U-SFAN} consists of self-learning methods that address the issue of unreliable model predictions. 
Uncertainty estimation has its foundation in either model uncertainty~\cite{CoWA-JMDS, U-SFAN} or data uncertainty~\cite{NRC, CoWA-JMDS}. Specifically, NRC~\cite{NRC} promotes label consistency by attributing neighbor affinity to loss weights based on proximity to nearest neighbors.
Among pseudo-labeling approaches, SHOT~\cite{SHOT} generates pseudo labels by clustering target features, a technique also adopted by CPGA~\cite{CPGA} and CDCL~\cite{CDCL}. Both CDCL and CPGA seek alignment of target features to source prototypes in accordance with their pseudo labels.
The issue of noisy pseudo labels arising from domain discrepancies is a recognized challenge addressed by methodologies like D-MCD~\cite{D-MCD}, VDM-DA~\cite{VDM-DA}, and SSNLL~\cite{SSNLL}. D-MCD employs a dual-classifier framework for adversarial training. By utilizing the CDD~\cite{BCDM} distance metric between the classifiers, it identifies and subsequently mitigates pseudo-label noise in high-confidence samples through a co-training process with a model pre-trained on ImageNet. VDM-DA employs a domain discriminator to harmonize the virtual and target domains, refining the model by selecting clean pseudo-labels based on sample entropy values. SSNLL employs an exponential-momentum-average model to segment the dataset, generating pseudo-labels for high-confidence samples while refining the label noise for those with lower confidence.
Unlike these methods that necessitate additional models for noise reduction, our approach uniquely identifies clean samples by exclusively examining the feature space of a singular model, thereby sidestepping potential pitfalls associated with additional modeling errors.

\subsection{Learning with Noisy Labels}
Deep neural networks (DNN) often exhibit a memorization effect, where they initially learn easy patterns but eventually overfit to noisy samples. This phenomenon has been observed in various deep networks~\cite{LNLC, RL-NL}.
The problem of learning with noisy labels aims to train a robust deep network using a dataset with noisy labels.
Sample selection methods have been proposed to filter out noisy samples based on the learning properties of DNN, such as loss distribution analysis~\cite{DivideMix}.
This intrinsic motivation serves as the driving force behind our proposed adaptive sample selection module, which aims to select clean pseudo labels for improved SFUDA performance.

More recently, Supervised Contrastive Learning~\cite{SupCon} (SCL) has also been employed for learning with noisy labels~\cite{MOIT, Sel-CL}.
MOIT~\cite{MOIT} combines SCL with mixup data augmentation~\cite{mixup} to prevent the memorization of label noise.
Sel-CL~\cite{Sel-CL} addresses the degradation issue in SCL by selecting confident samples.
Considering that pseudo labels in SFUDA are also subject to noise due to the domain gap between source and target data, reducing label noise becomes crucial.
However, Sel-CL cannot be directly applied to the SFUDA task as the labels of target data are inaccessible.
Therefore, compared to Sel-CL, our method relies more on the predictions of the source model rather than training data information and is simpler as it does not require a queue or the mixup technique.

\section{Method}
In this section, we first describe the problem setting for SFUDA. Then we introduce our proposed \modelname{} (\shortname) framework. Firstly, \selectionname{} (\selectshortname) filters out noisy pseudo labels by simply analyzing the information from neighboring samples. This process divides the target data into a clean pseudo-labeled set and a noisy pseudo-labeled set. Subsequently, \clname{} (\clshortname) is applied to the clean pseudo-labeled set to further enhance the model's robustness against pseudo-label noise.
The filtering occurs 
at each epoch, allowing the confident samples to be dynamically updated.

\begin{figure*}[t]
\centering
\includegraphics[width=0.7\textwidth]{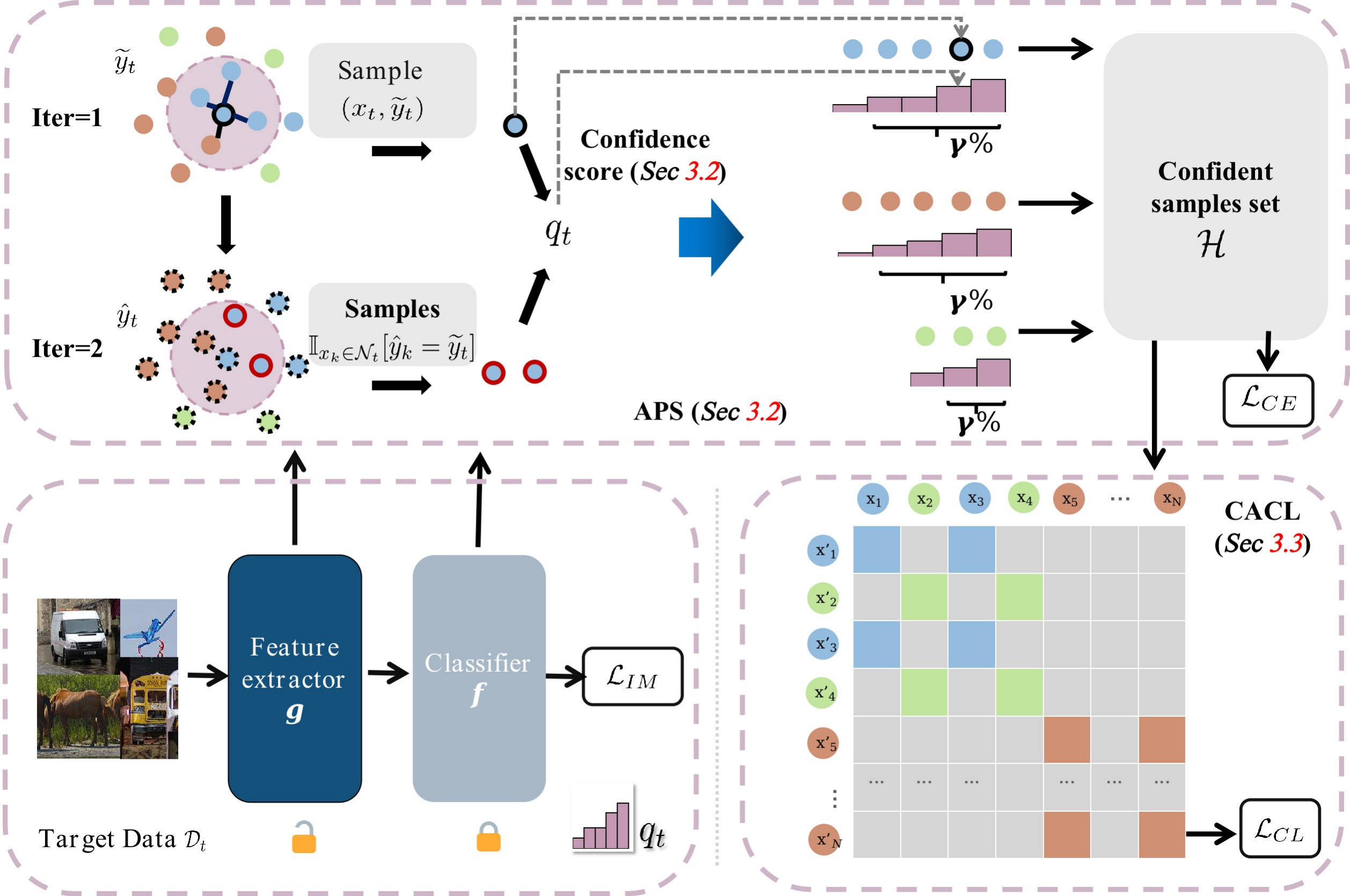}
\caption{
A comprehensive framework of \shortname{} for addressing SFUDA task. 
$\mathcal L_{IM}$ is applied to all samples.
The \selectshortname{} module performs two rounds of uncertainty estimation and samples $\mathbb I_{x_k\in\mathcal N_t}[\hat y_k = \widetilde y_t]$ are useful for $q_t$. 
$q_t$ is ranked within each class to form a confident sample set $\mathcal H$, and then apply $\mathcal L_{CE}$ and $\mathcal L_{CL}$ to $\mathcal{H}$. Specifically, each sample $x_i$ and
its strongly augmented view $x_i'$ are used in the class-wise contrast (represented by the colored matrix block) of $\mathcal L_{CL}$. 
}
\label{fig:structure}
\end{figure*}

To provide a visual overview of our methodology, we present Figure~\ref{fig:structure} to illustrate the entire framework.

\subsection{Preliminaries}

\mytext{Notation.} This work aims to address the SFUDA task for image classification.
We denote the labeled source dataset as $\mathcal{D}_s = \{(x_s, y_s)\}$, where $x_s\in \mathcal{X}_s$ and $y_s\in \mathcal{Y}_s$.
The target data is represented as $\mathcal{D}_t$ and consists of only unlabeled images $\{x_t\}$, where $x_t\in \mathcal{X}_t$.
The given model $\phi$ consists of a feature extractor $g:\mathcal{X} \rightarrow \mathbb{R}^{D}$, followed by a classifier $f:\mathbb{R}^{D} \rightarrow \mathbb{R}^{C}$, where $D$ denotes the dimensionality of the feature space and $C$ represents the number of classes in $\mathcal{D}_s$ and $\mathcal{D}_t$.

In the context of SFUDA, the given model $\phi$ is pre-trained on $\mathcal{D}_s$, and $\mathcal{D}_s$ cannot be used during adaptation. Due to domain shifts, the model gives uncertain predictions on the target data, and noise exists in the pseudo labels.
Our objective is to progressively select clean pseudo labels and make the model $\phi$ adapt to the target data $\mathcal{D}_t$. We freeze the classifier $f$ and update the feature extractor $g$ during the entire adaptation process.

\mytext{Pseudo-Labeling.}
Pseudo labels, as introduced in SHOT~\cite{SHOT}, are used in our method and need to be denoised, as shown in Fig.~\ref{fig:fig1}. In general, SHOT estimates class feature distributions weighted by both the model's prediction probability and the class sample count.

In detail, the process of pseudo-labeling first aggregates features based on the prediction probability assigned to the target training samples. The class feature aggregation is performed as follows:
\begin{equation}
\mu_c^{(0)}=\frac{\sum_{{x_t}\in \mathcal{X}_t}\delta_c(\phi(x_t))g(x_t)}{\sum_{{x_t}\in \mathcal{X}_t}\delta_c(\phi_t(x_t))}, \quad c \in [1,\cdots,C].
\label{con:mu_0}
\end{equation}

Here, $\delta$ represents the softmax function, with its $c$-th element denoted as $\delta_c$.
Then, the intermediate pseudo labels $\widetilde {y}^{(0)}_t$ are obtained by assigning them to the nearest class center (according to the cosine similarity metric $d(,)$) in the feature representation space, as follows:

\begin{equation}
\widetilde {y}_t^{(0)}=\arg\max_c d(g(x_t),\mu^{(0)}_c)
\label{con:y_0}
\end{equation}

Finally, the class centroids $\mu^{(1)}_c$ are weighted by the class sample count, resulting in the acquisition of the final pseudo labels $\widetilde y_i^{(1)}$.
\begin{align}
    &\mu^{(1)}_c=\frac{\sum_{{x_t}\in \mathcal{X}_t}\mathds{1}(\widetilde{ y}_t^{(0)}=c)g(x_t)}{\sum_{{x_t}\in \mathcal{X}_t}\mathds{1}(\widetilde{y}_t^{(0)}=c)} \notag \\
    &\widetilde{y}_t^{(1)}=\arg\max_c d(g(x_t),\mu^{(1)}_c) \label{con:mu_1_y_1}
\end{align}
Here, the indicator function $\mathds 1()$ equals 1 if $\widetilde y_t^{(0)}$ equals c, and 0 otherwise. We will use $\widetilde y_t$ to replace $\widetilde y_t^{(1)}$ in the following text.

\begin{figure}[t]
\centering
\includegraphics[width=0.7\columnwidth]{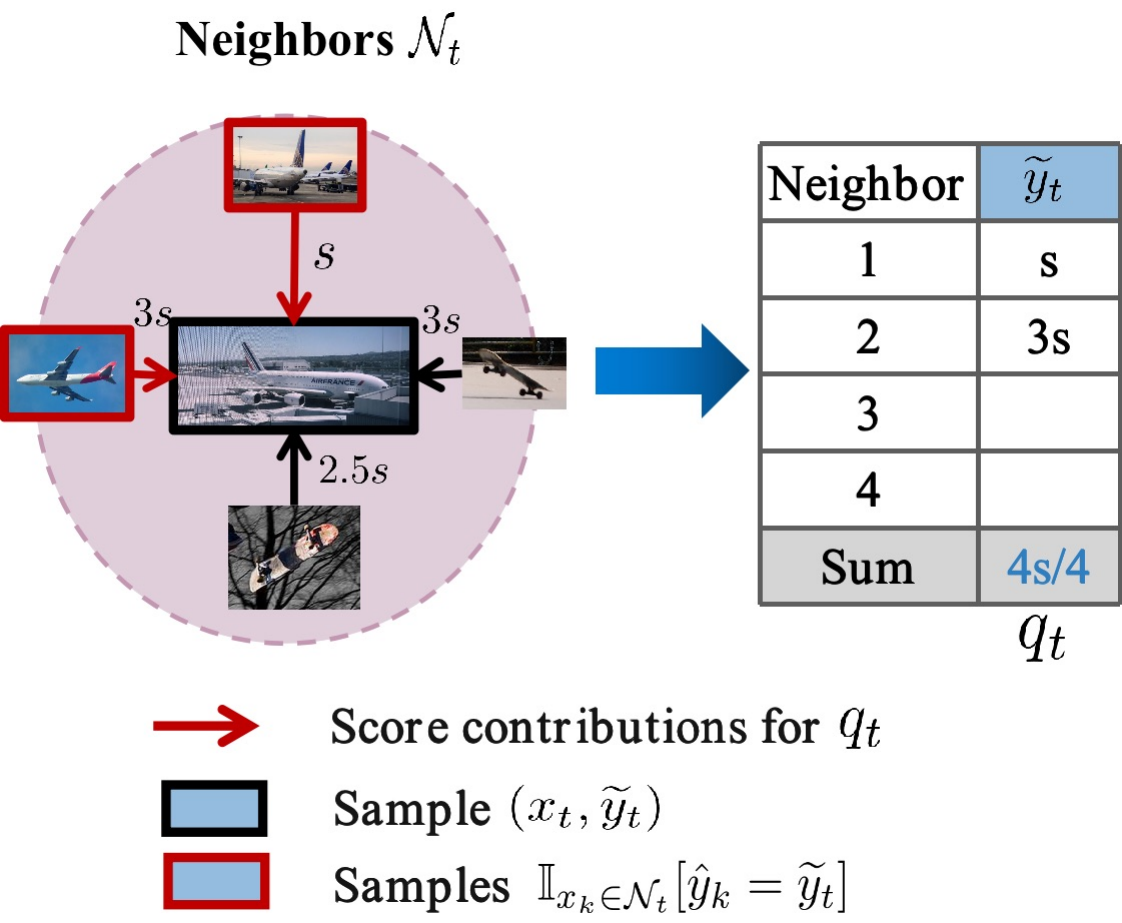} 
\caption{
The calculation of confidence score $q_t$ with cosine similarity $s$.
}
\label{fig:confidence_score}
\end{figure}

\subsection{\selectionname}

The issue of pseudo-label noise arises when considering the overall data representation distribution for pseudo-label generation. Our hypothesis is that easily clustered samples in the feature space $g(x_t)$ contain a higher degree of domain-shared information due to the presence of similar features within the same class, while numerous uncertain samples reside between cluster centers, primarily caused by distribution shifts between domains. Inspired by this, we propose a novel pseudo-label selection mechanism, \selectshortname, that iteratively estimates the uncertainty of samples via nearby neighbors.

In detail, for each sample $x_t$, 
\selectshortname{} construct its $K$-nearest neighbors $\mathcal{N}_t$ based on the pairwise cosine distance in the feature space. Subsequently, 
neighbors $\mathcal{N}_t$ are leveraged to perform voting based on distance and estimate the class posterior probabilities $\hat p_t \in \mathbb{R}^C$ for sample $x_t$.

\begin{equation}
\hat p_t=\frac{1}{K} \sum_{\substack{ k=1\\x_k\in \mathcal{N}_t}}^K {d(g(x_t),g(x_k))} \bm{e}_{\hat y_k}
\label{con:confidence}
\end{equation}

Here, $\bm{e}_{\hat y_k} \in \mathbb{R}^C$ represents the one-hot vector of the temporary pseudo-label $\hat y_k$. $\hat y_k$ is initialized with $\widetilde y_k$ and is updated during the iteration. The temporary pseudo labels $\hat y_t$ for all target samples are updated using the following equation:

\begin{equation}
\hat y_t=\arg\max_c \hat p_{t}^{c}
\label{con:maxl}
\end{equation}
Where the superscript $c$ indicates taking a value from the $c$-th class.

The estimation of $\hat p_t$ is iteratively executed by updating $\hat y_t$ to sense the density of the surrounding pseudo labels and obtain a more accurate uncertainty estimation. Through empirical observation, it has been determined that two iterations of this estimation process (Eq.~\eqref{con:confidence} $\to$ Eq.~\eqref{con:maxl} $\to$ Eq.~\eqref{con:confidence}) are adequate for convergence.
Once the estimation is finalized, the confidence score $q_t$ is defined as follows: 
\begin{equation}
    q_t =\frac{1}{K} \sum_{\substack{ k=1\\x_k\in \mathcal{N}_t}}^K \mathds{1}(\hat y_k=\widetilde y_t){d({g(x_t),g(x_k)})}
    \label{con:q}
\end{equation}
Where, for sample $x_t$, only those nearest neighbors that have the same label as $\widetilde y_t$ in the end, denoted as $\mathbb{I}_{x_k \in \mathcal N_t}[\hat y_k =\widetilde y_t]$, will contribute to $q_t$, as shown in Fig.~\ref{fig:confidence_score}.

\mytext{Pseudo-label Selection Criterion.}
Based on the model prediction uncertainty hypothesis and to avoid omitting categories in the pseudo-label selection, 
samples within each pseudo-class are first sorted based on their confidence score $q_t$ at every epoch. Then, 
the top $\gamma$\% of confident samples for each pseudo-class are selected.
These selected samples are then grouped into a confident sample set $\mathcal{H}$.

\begin{figure}[t]
\centering
\includegraphics[width=0.9\columnwidth]{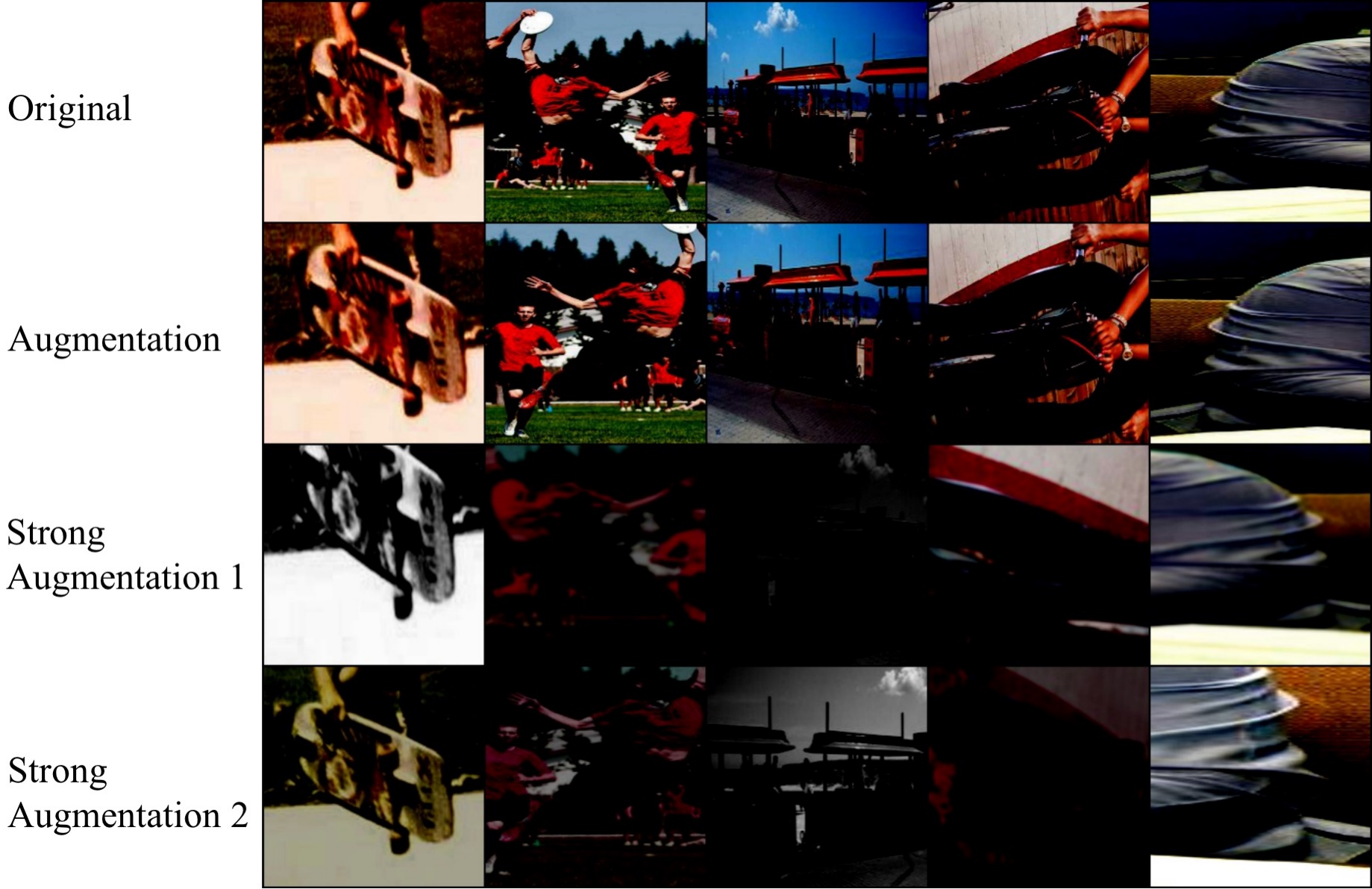}
\caption{
Enhancing data diversity through augmentation techniques: an illustration.
}
\label{fig:augmentation}
\end{figure}

\subsection{\clname}
Inspired by previous studies~\cite{MOIT, Sel-CL} on label noise learning, we aim to enhance semantic learning and mitigate the impact of label noise through the incorporation of category-wise contrastive learning. Hence, we employ SCL~\cite{SupCon} on the confident sample set $\mathcal{H}$ with their pseudo labels, where positive pairs are formed from multiple image views within the same class. 

\mytext{Contrastive Loss.}
Concretely, within a mini-batch of size $B$, we define $i \in I=\{1, \cdots, 2B\}$ as the index of an arbitrary augmented training sample. The strong augmented views, as illustrated in Fig.~\ref{fig:augmentation}, are considered. To create positive pairs, 
the selected confident samples are exhaustively coupled if they share the same pseudo-class.
Formally, positive pair set for anchor $i$ is formed as:
{
\begin{equation}
    \mathcal{O}(i)=\{o_{ij}|y_i=y_j ; (x_i,y_i),(x_j,y_j)\in \mathcal{H}\} 
\label{con:class_pair}
\end{equation}}
where $o_{ij}=<(x_i,y_i),(x_j,y_j)>$ and $j \in A(i)$ where $A(i)\equiv I \backslash\{i\}$.

Then, the class-wise contrastive loss is formulated as:
\begin{equation}
    \mathcal{L}_{CL}=\sum_{i\in I}\frac{-1}{ \left| \mathcal{J}(i) \right|} \sum_{j\in \mathcal J(i)} \log \frac{\exp(z_i \cdot z_j/\tau)}{\sum _{a\in A(i)} \exp(z_i \cdot z_a/\tau)}
\label{con:loss_cl}
\end{equation}
Here, $\mathcal{J}(i)=\{j|j\in A(i),o_{ij} \in \mathcal{O}(i) \}$, 
$\tau$ is the temperature parameter, 
and $z_{i(j)}=\frac{\phi(x_{i(j)})}{\left\|\phi(x_{i(j)}) \right\| }$ is the normalized feature representation.

\mytext{Cross-Entropy Loss.}
We apply the standard cross-entropy loss on the confident sample set $\mathcal{H}$ to guide the adaptation until convergence, and we denote the loss as $\mathcal{L}_{CE}$.

\begin{equation}
    \mathcal{L}_{CE} = -\mathbb E_{(x_t,\widetilde y_t)\in \mathcal{H} } \sum_{c=1}^{C}\mathds{1}(c=\widetilde y_t) log \delta_c(\phi(x_t))
\end{equation}

\mytext{Information Maximization Loss.}
The information maximization loss $\mathcal{L}_{IM}$~\cite{SHOT} aims to extract and utilize the relevant information from the input data and cluster all the target data.

Finally, the overall loss function is presented as :
\begin{equation}
    \mathcal{L}_{ALL}= \mathcal{L}_{CL}+\beta\mathcal{L}_{CE} + \mathcal{L}_{IM}
    \label{con: all loss}
\end{equation}
where $\beta>0$ serves as a balancing factor. 

\subsection{Progressive pseudo-label selection}

\begin{algorithm}[ht]
\caption{ Training procedure for \shortname{}}
\label{alg:algorithm}
\textbf{Input}: Unlabeled target samples $\mathcal D_t$ and
a pre-trained source model $\phi$.

\textbf{Output}: An adapted model $\phi$ for the target domain.

\textbf{Parameters}: The epochs $\hat T$, iterations (mini-batch number) $\widetilde T$ per epoch, batch size $B$;
\begin{algorithmic}[1] 

\FOR {$epoch=1$ \TO $\hat T$}
\STATE Obtain pseudo labels via Eq.~\eqref{con:mu_0} - Eq.~\eqref{con:mu_1_y_1}.
\STATE Compute confidence score via Eq.~\eqref{con:confidence} - Eq.~\eqref{con:q} and select the top $\gamma$\% confident ones to build sample set $\mathcal{H}$.
\FOR {$iter=1$ \TO $\widetilde T$}
\STATE Sample mini-batch $\{x_b\}_{b=1}^{B}$ in $\mathcal{D}_t$ and get their corresponding pseudo labels.
\STATE Get positive pair sets $O$ via Eq.~\eqref{con:class_pair}.
\STATE Update model $\phi$ via $\mathcal{L}_{ALL}$ in Eq.~\eqref{con: all loss}.
\ENDFOR
\ENDFOR
\end{algorithmic}
\end{algorithm}

Given the substantial domain shifts that often exist between the source and target domains, it is impractical to accurately determine the complete set of true pseudo labels in a single iteration. Therefore, we adopt a progressive strategy that alternates between two steps: (1) selecting the clean pseudo-labeled samples, and (2) adapting the model accordingly. This approach aligns well with the epoch-based protocol commonly employed in deep learning. The complete training procedure for \shortname{} is outlined in Algorithm \ref{alg:algorithm}.

\section{Implementation.}
\mytext{Datasets.}
We evaluate our method on three widely used SFUDA datasets: Office, Office-Home, VisDA-C and DomainNet-126. (1) {\bf Office}~\cite{office-31} is a small-sized dataset that contains three domains (Amazon (A), Dslr (D) and Webcam (W)) in the real office scenarios, each of which has 31 classes. 
(2) {\bf Office-Home}~\cite{office-home} is a media-sized dataset with 65 classes across four domains: Art (Ar), Clipart (Cl), Product (Pr) and Real-World (Rw). 
(3) {\bf VisDA-C}~\cite{Visda-c} is a large-sized dataset with 12 classes, with the adaption task from source synthetic (S) domain to target real (R) domain. 
(4) \textbf{DomainNet-126} is a subset of DomainNet~\cite{DomainNet} used by previous work~\cite{MME}. It has 126 classes from 4 domains (Real, Sketch, Clipart, Painting) and we evaluate 7 domain shifts built from the 4 domains.

\mytext{Baselines.}
For source-available unsupervised domain adaptation, we compare our proposed methods against a number of baselines: CDAN~\cite{CDAN}, MCC~\cite{MCC}, MDD~\cite{MDD}, BCDM~\cite{BCDM}, FixBi~\cite{FIxBi}, CAN~\cite{CAN} and SRDC~\cite{SRDC}.
For source-free unsupervised domain adaptation, we compare with these baselines: 3C-GAN~\cite{3C-GAN}, U-SFAN~\cite{U-SFAN}, DIPE~\cite{DIPE}, HCL~\cite{HCL}, NRC~\cite{NRC}, CoWA-JMDS~\cite{CoWA-JMDS}, AdaContrast~\cite{AdaContrast}, CDCL~\cite{CDCL}, Tent~\cite{TENT}, D-MCD~\cite{D-MCD}, VDM-DA~\cite{VDM-DA}, SSNLL~\cite{SSNLL} and SHOT~\cite{SHOT}. 

\mytext{Pretraining on Source Domain.}
To ensure a comprehensive comparison with previous methods~\cite{SHOT, CoWA-JMDS, CDCL, U-SFAN, DIPE, NRC, VDM-DA, SSNLL}, 
we utilize the same ResNet~\cite{Resnet} models. Specially, we use ResNet-50 for Office, Office-Home and DomainNet-126, ResNet101 for Visda-C.
The ResNet's feature extractor component is adopted as the backbone, while the final fully connected (FC) layer is substituted with a bottleneck layer and a task-specific FC layer.

To enhance the model's performance, batch normalization is introduced after the FC in the bottleneck layer, and weight normalization is applied prior to the FC in the last task-specific FC layer. These additional layers contribute to improved learning and generalization capabilities.
We denote the combined backbone and bottleneck layer as the feature extractor denoted by $g$, while the last FC layer is referred to as the classifier denoted by $f$. The feature dimension is fixed at $D=256$, while the output dimension of the model is denoted by $C$.

During source training, we adopt the label smoothing loss~\cite{LabelSmooth} and partition the source dataset into a 9:1 train-to-validation ratio, following the approach outlined in~\cite{SHOT}. We employ different epochs for different datasets, namely 100 for Office, 50 for Office-Home, 30 for DomainNet-126 and 10 for VisDA-C. 

\mytext{Adaptation on Target Domain.}
The target model is initialized using the pre-trained source model, and the classifier weights remain fixed throughout the SFUDA stage. A stochastic gradient descent optimizer is employed with a momentum of 0.9 and a weight decay of 1e-3. The training batch size is set to 64. For the Office, Office-Home and DomainNet-126 datasets, the initial learning rate $\eta_0$ is configured as 1e-2, whereas for VisDA-C, it is set to 1e-3.

In the training process, the learning rate of the bottleneck layer is determined to be 10 times that of the backbone. The training procedure spans 15 epochs, starting with a linear learning rate warmup exclusively during the first epoch. After the warmup phase, the learning rate gradually decreases according to the cosine decay schedule, without any restarts.

To ensure convergence during training, we empirically set the loss weights as $\beta=0.3$. Furthermore, the following hyperparameters are defined: temperature 
$\tau=0.1$, neighborhood size $K=4$, and the ratio $\gamma=0.6/0.6/0.6/0.8$ for the Office, Office-Home, DomainNet-126, and VisDA-C datasets, respectively. Data augmentation in contrastive learning adheres to the same settings as MoCov2~\cite{MOCOV2}.

\input{tables/sota_visda}
\input{tables/sota_office-home}

\input{tables/sota_office}
\input{tables/sota_domainnet}

\section{Experiments}
\subsection{Experimental Results.}

\mytext{VisDA-C Results.} 
As shown in Table \ref{table:Visda-C}, our method achieves a significant improvement over the base method SHOT (82.9\%) with a substantial margin of 5.8\%. This highlights the effectiveness of our approach in filtering pseudo-label noise and preventing label noise memorization, thereby enhancing the performance of SFUDA.
Furthermore, our method surpasses the previous state-of-the-art SFUDA methods by a remarkable margin of 1.2\%. In comparison to other UDA methods, 
a notable improvement of 1.5\% is achieved. Notably, when considering intra-class transfer tasks, our \shortname{} 
achieves state-of-the-art performance in terms of average accuracy, demonstrating its superiority over previous solutions.

It is important to mention that several existing methods: D-MCD~\cite{D-MCD} (87.5\%), VDM-DA~\cite{VDM-DA}, and SSNLL~\cite{SSNLL}, also tackle the issue of denoising pseudo labels using specific criteria. Nonetheless, our \shortname{} outperforms these methods, demonstrating substantial improvements in comparison.

\mytext{Office-Home Results.}
Table~\ref{table:office-home} displays the experimental results on the Office-Home dataset. Our method exhibits a significant improvement, achieving a substantial margin of 1.2\% over the SHOT baseline. It is noteworthy that our approach outperforms existing state-of-the-art SFUDA and UDA methods, establishing a new benchmark. Specifically, 
the highest accuracy across 5 domain shift tasks are achieved, including Cl$\to$Pr and Cl$\to$Rw.

\mytext{Office Results.}
The results on the Office dataset are presented in Table~\ref{table:office}. Despite the limited size of the training dataset, our method demonstrates comparable performance to the state-of-art SFUDA methods. Specifically, we observe a significant improvement of over 1.3\% compared to the baseline SHOT approach.

\mytext{Domainnet-126 Results.}
Table~\ref{table:domainnet} reports the results of 7 domain shift tasks. Among these methods, our UPA, which follows the SFUDA approach, stands out with an impressive average accuracy of 68.0\%.  When compared with other methods, UPA not only surpasses the baseline SHOT by a margin of 0.9\% but also outperforms other competitive methods.

\input{tables/ablation_loss.tex}
\input{tables/ablation_pseudo}
\input{tables/ablation_pe}

\subsection{Ablation Studies}
We conduct a series of ablation study experiments to shed more light on the effectiveness of our method.

\mytext{Ablation of Submodules.}
We conducted a comprehensive series of ablation study experiments to further investigate the effectiveness of our proposed method. In Table~\ref{table:ablation_loss}, we present the results of our sub-components ablation study on three datasets. Specifically, we compare our method (Methods.3) with two other variants: Methods.1, which refers to the SHOT method, and Methods.2, which includes the \selectshortname{} module where the cross-entropy loss ($\mathcal{L}_{CE}$) is applied only to the confident sample set $\mathcal{H}$. 

When considering \selectshortname{} alone, we observe noticeable improvements of 0.9\%, 0.9\%, and 1.3\% on the Office, Office-Home, and VisDA-C datasets, respectively. This highlights the presence of pseudo-label noise in the SFUDA task, which can be effectively detected by the \selectshortname{} module. Furthermore, by incorporating contrastive learning to mitigate the impact of noisy label memorization, 
our method achieves additional performance improvements of 0.4\% and 0.3\% on the Office and Office-Home datasets, respectively. Notably, the \clshortname{} module demonstrates a significant performance boost of 4.5\% on the VisDA-C dataset. These results provide strong evidence for the efficacy of our proposed method and its ability to address the challenges of the SFUDA task.

\mytext{Effectiveness of Pseudo-label Selection.}
An ablation study was conducted to validate the effectiveness of the pseudo-label uncertainty estimation method within our \selectshortname{} module. Table~\ref{table:PL} presents the comparison of three different approaches for estimating pseudo-label uncertainty: (1) model probability $\phi(x_t)$ associated with the corresponding pseudo label (Prob), (2) negative entropy (Ent) of the prediction $\phi(x_t)$, and (3) cosine similarity (Cossim) with the clustering center $d(g(x_t), \mu^{(1)})$. The results demonstrate that our \selectshortname{} module achieves the highest performance on the transfer tasks S$\to$R of the VisDA-C dataset and Ar$\to$Pr task of the Office-Home dataset. This indicates that the uncertainty estimation using KNN neighbors outperforms the other methods.

\mytext{Iterative Refinement of \selectshortname.}
The iterative refinement of uncertainty estimation in \selectshortname{} aims to improve the accuracy of pseudo-label distribution perception. To validate our iterative design, we conducted experiments and presented the results in Table~\ref{tab:selection_times}. The table demonstrates that incorporating a second iteration leads to a significant improvement of 1.5\% on the VisDA-C dataset and 0.4\% on the Office dataset. However, additional iterations do not result in further improvements, suggesting that two iterations provide the optimal number for confidence score estimation.

\begin{figure*}[ht]
\centering
\begin{subfigure}[b]{0.3\textwidth}
    {
    \centering
    \includegraphics[width=\textwidth]{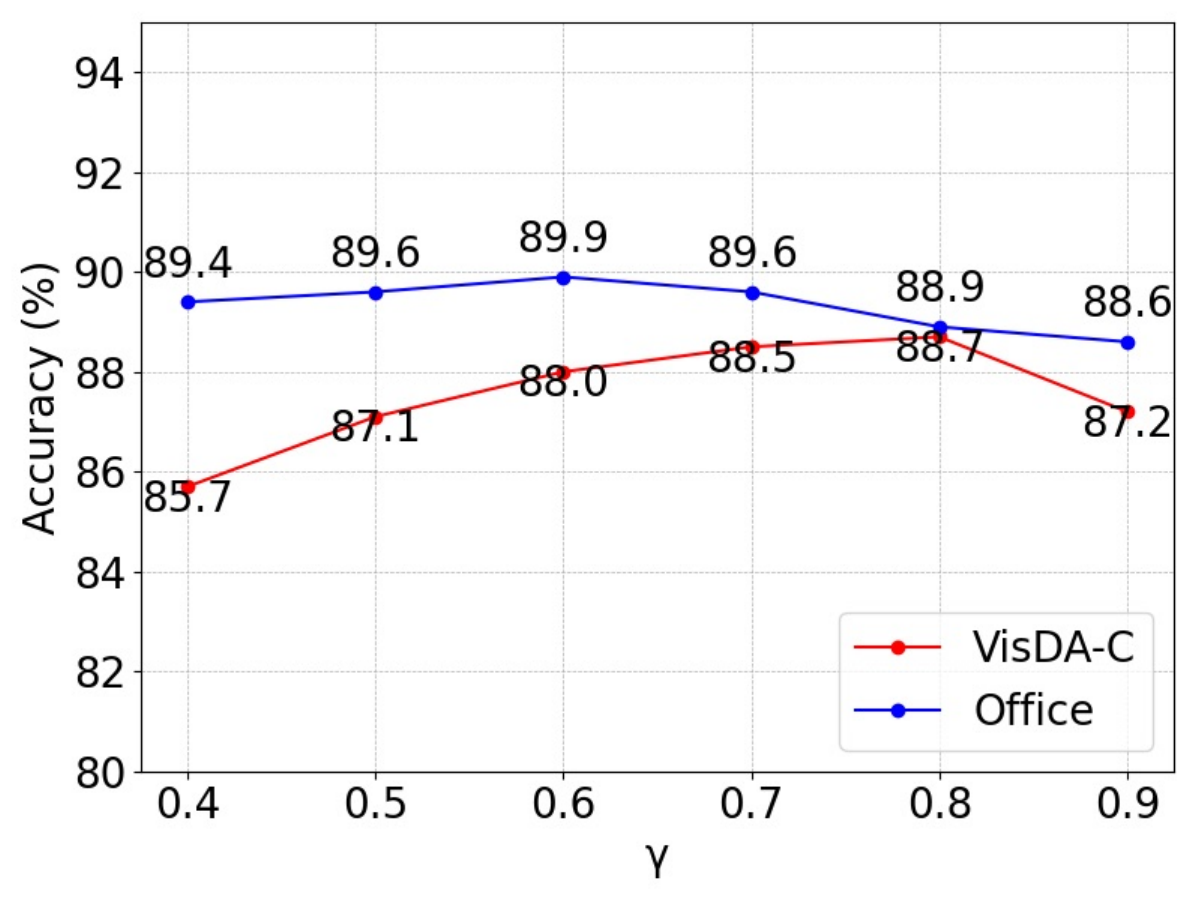}
    \caption{}
    \label{fig:param-a}
}
\end{subfigure}
\begin{subfigure}[b]{0.3\textwidth}
    {
    \centering
    \includegraphics[width=\textwidth]{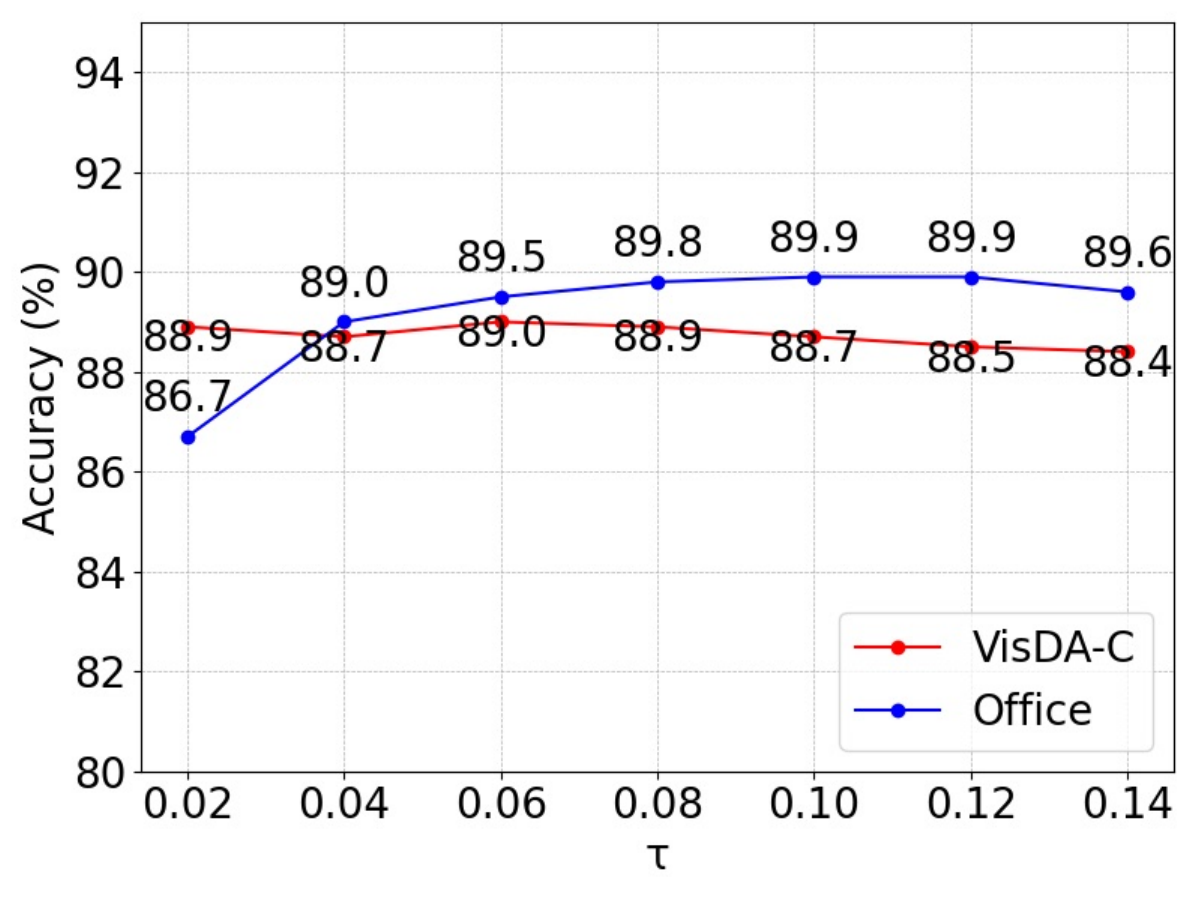}
    \caption{}
    \label{fig:param-b}
}
\end{subfigure}
\begin{subfigure}[b]{0.3\textwidth}
    {
    \centering
    \includegraphics[width=\textwidth]{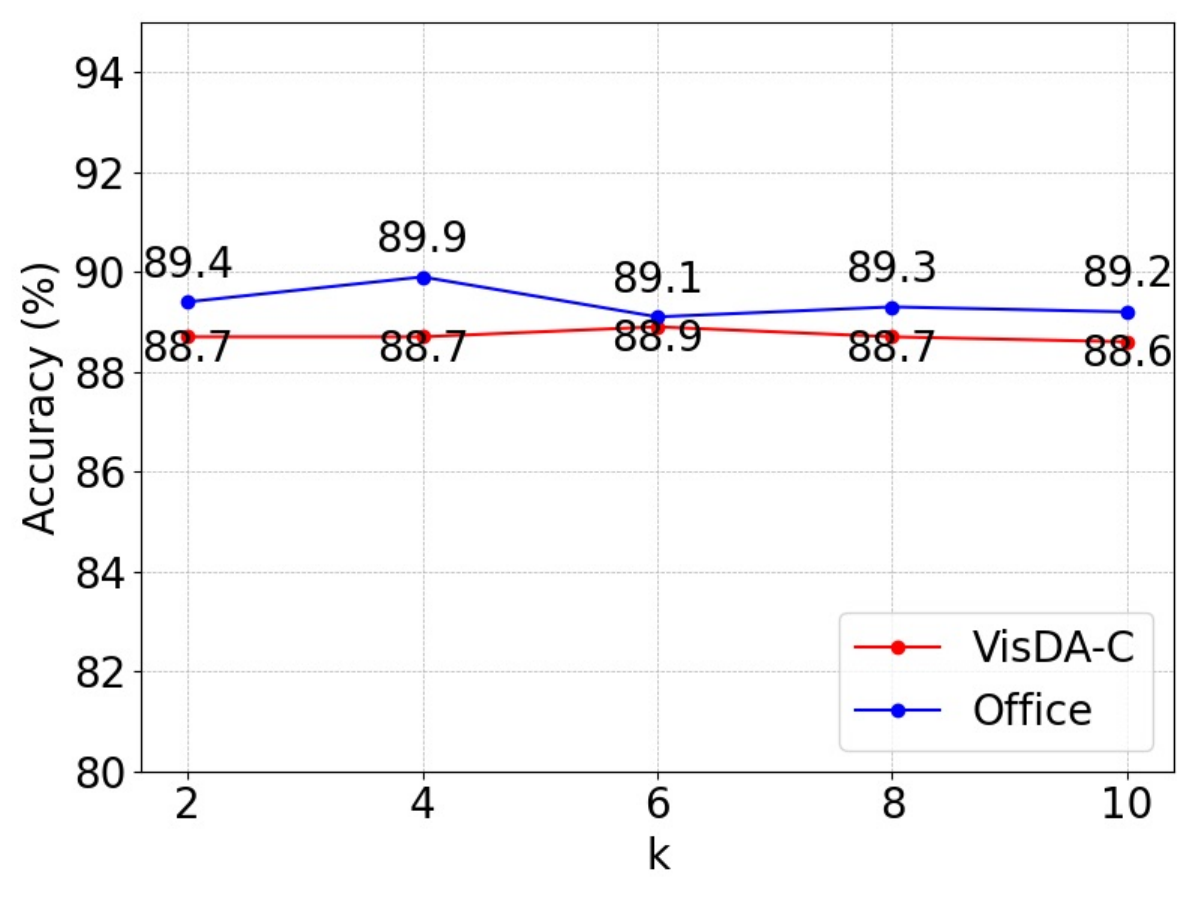}
    \caption{}
    \label{fig:k}
}
\end{subfigure}
\caption{
    Parameter sensitivity analysis for SFUDA tasks using \shortname{} on VisDA-C and Office datasets. 
    (a) The confident sample selection ratio $\gamma$;
    (b) The temperature $\tau$ in $\mathcal{L}_{CL}$;
    (c) The neighborhood size $K$.
    }
\label{fig:param}
\end{figure*}

\begin{figure*}[ht]
\centering
\begin{subfigure}[b]{0.2\textwidth}
    \includegraphics[width=\textwidth]{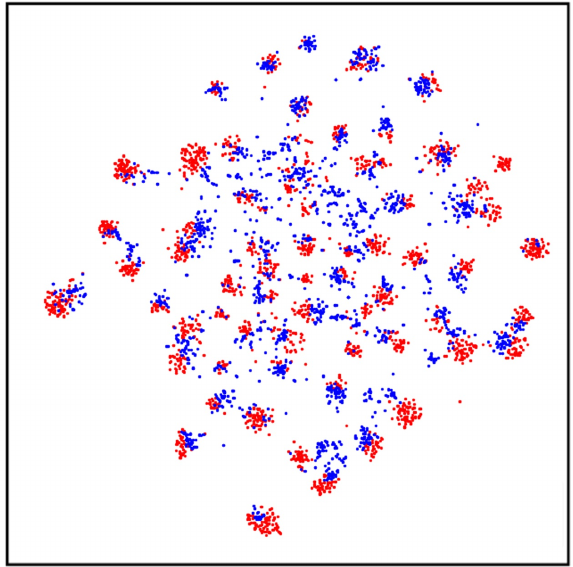}
    \caption{}
    \label{fig:tsne-a}
\end{subfigure}
\begin{subfigure}[b]{0.2\textwidth}
    \includegraphics[width=\textwidth]{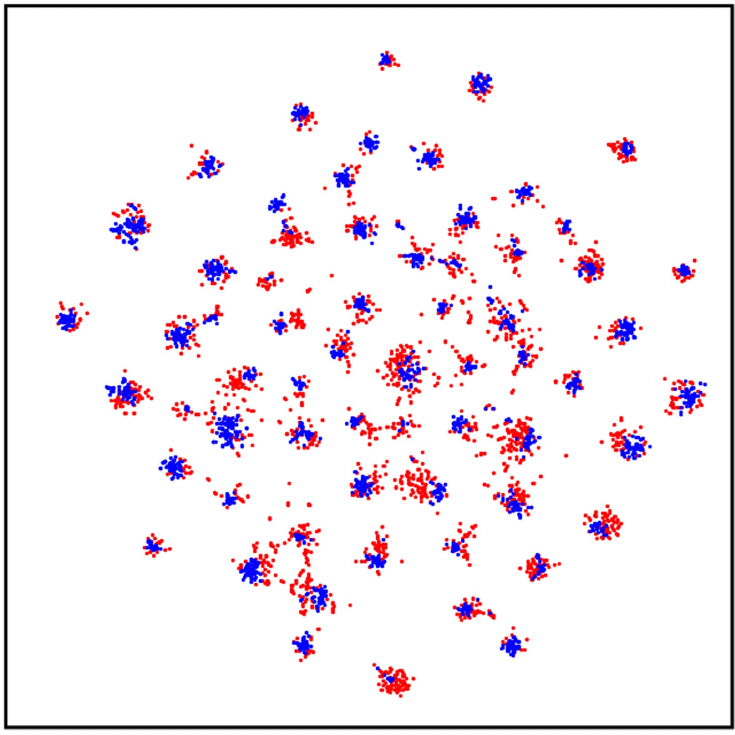}
    \caption{}
    \label{fig:tsne-b}
\end{subfigure}
\begin{subfigure}[b]{0.2\textwidth}
    \includegraphics[width=\textwidth]{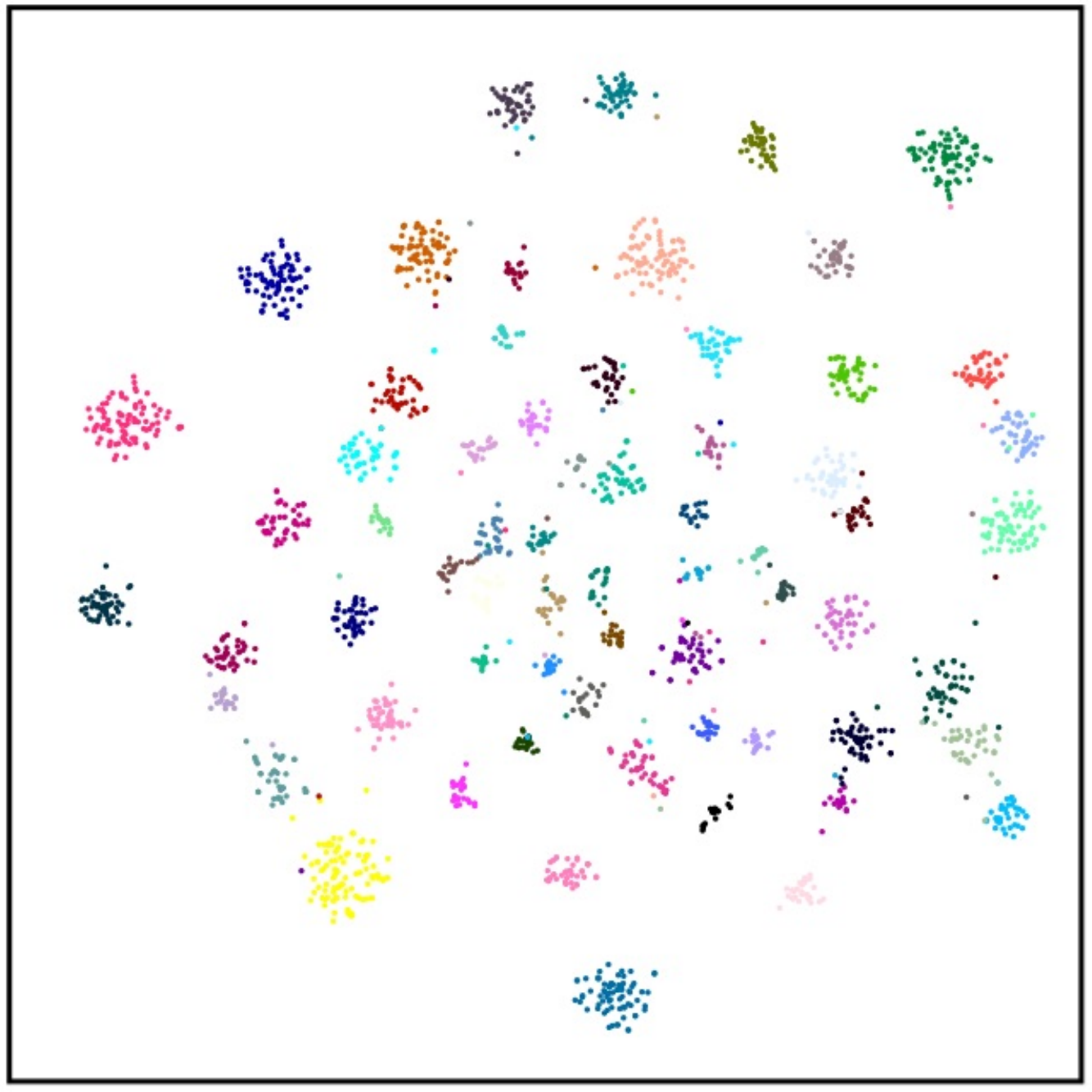}
    \caption{}
    \label{fig:tsne-c}
\end{subfigure}
\begin{subfigure}[b]{0.2\textwidth}
    \includegraphics[width=\textwidth]{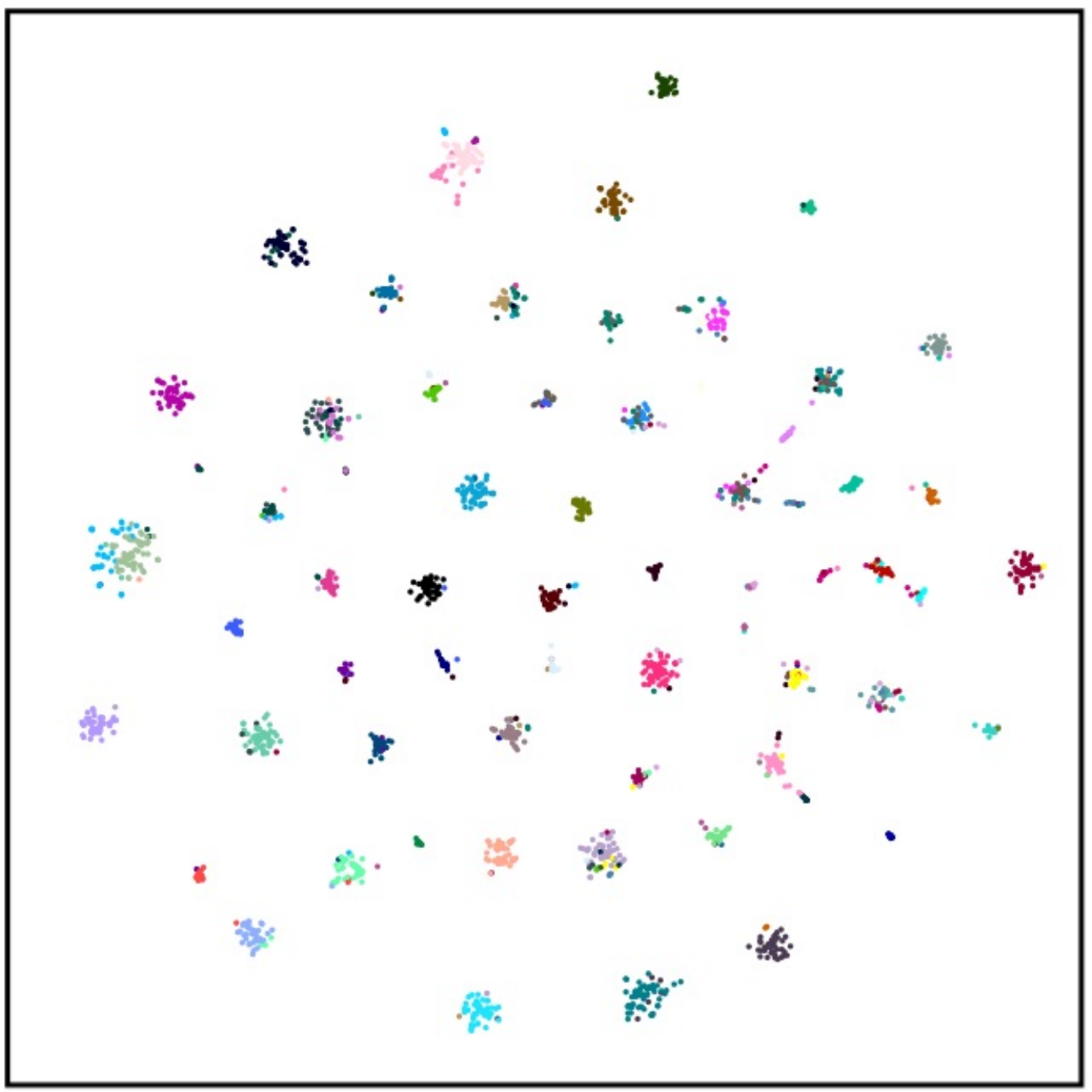}
    \caption{}
    \label{fig:tsne-d}
\end{subfigure}
\caption{
    The t-SNE visualization of SFUDA task Ar$\to$Pr (Office-Home) with ResNet50. (a)(b): the source features (i.e. red dots) and target features (i.e. blue dots) visualization before and after adaptation; (c)(d): the class-wise distribution (i.e. different colors) of target features before and after adaptation.
    }
\end{figure*}

\begin{figure}[ht]
    \centering
    \includegraphics[width=0.9\columnwidth]{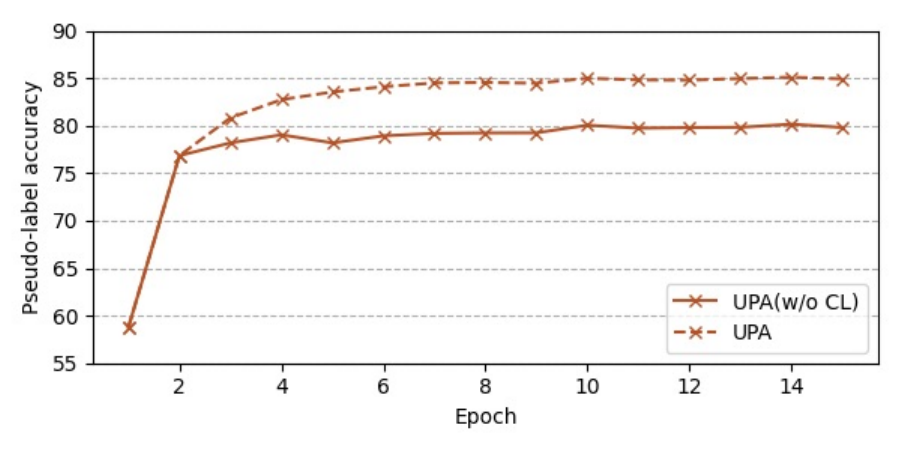}
    \caption{Evolution of selected pseudo-label accuracy on Visda-C dataset.}
    \label{fig:CL_acc}
\end{figure}
\begin{figure}[ht]
\centering
\begin{subfigure}[b]{0.45\columnwidth}
    {
    \includegraphics[width=\columnwidth]{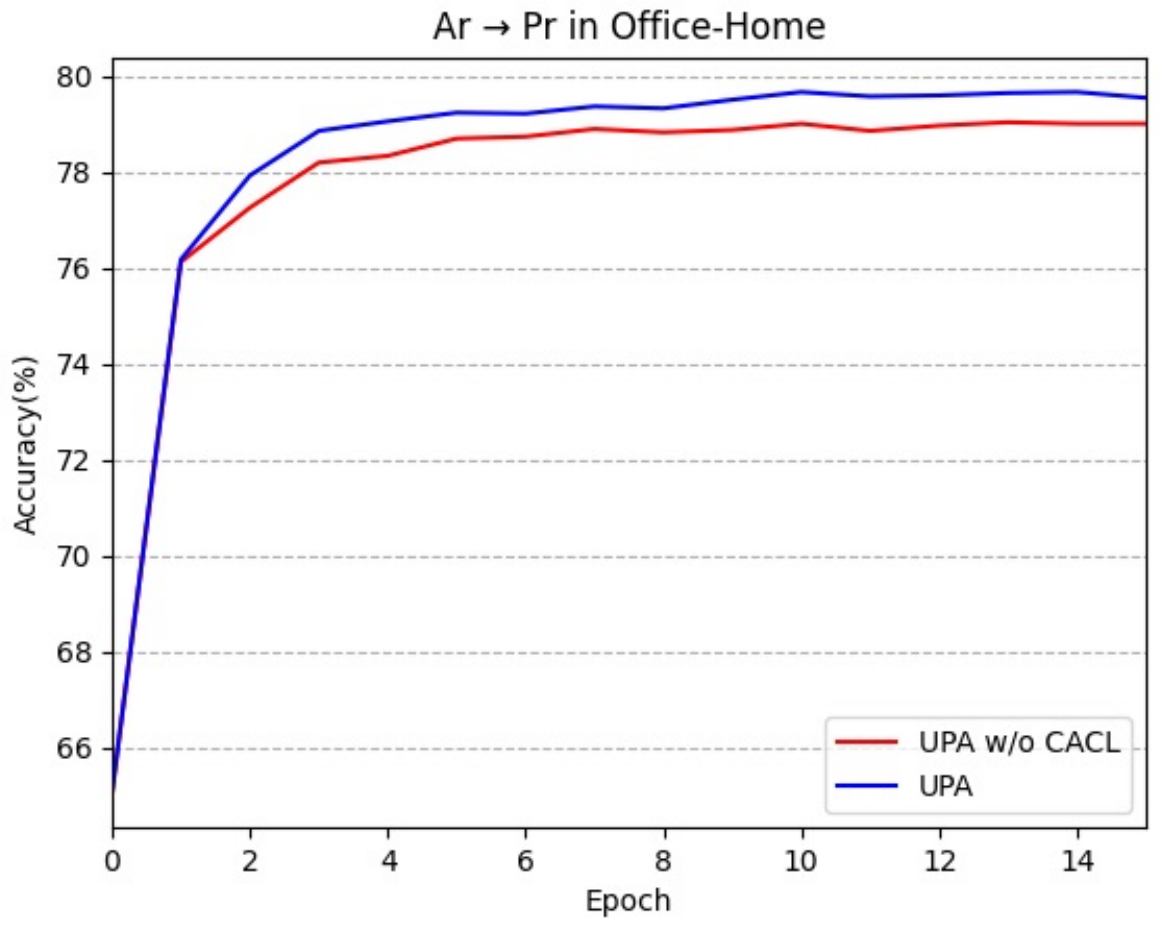}
    \caption{}
    \label{fig:ap_acc}
}
\end{subfigure}
\begin{subfigure}[b]{0.45\columnwidth}
    {
    \includegraphics[width=\columnwidth]{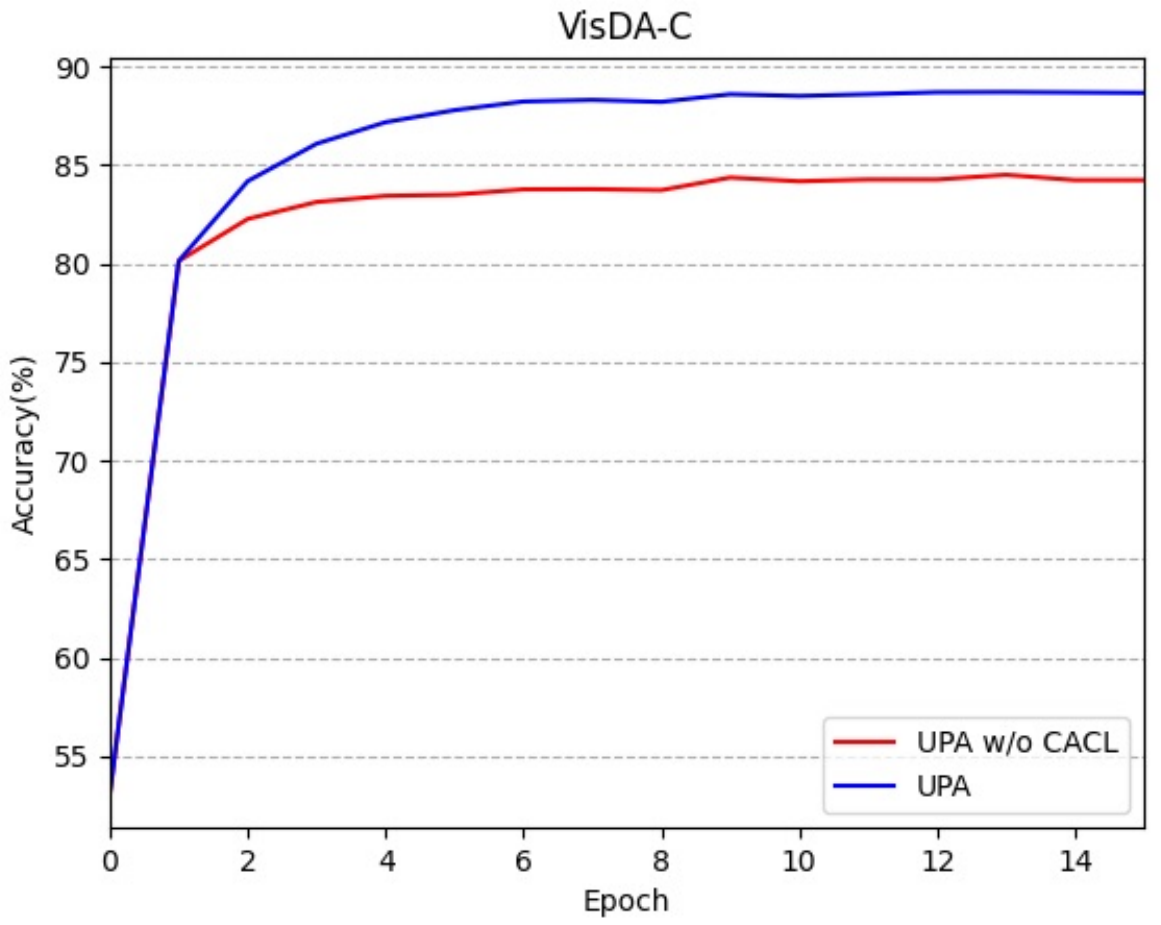}
    \caption{}
    \label{fig:vis_acc}
}
\end{subfigure}
\caption{
Accuracy Curves for SFUDA Tasks with \shortname{}(w/o \clshortname) and \shortname{}.
(a) Ar$\to$Pr (Office-Home) Task;
(b) S$\to$R (VisDA-C) Task.
}
\end{figure}

\mytext{Peventing Label Noise Memorization.}
The proposed method, \clshortname{}, incorporates class-wise contrastive learning~\cite{SupCon} by leveraging the selected clean pseudo labels. This integration aims to promote consistent probability outputs across diverse views of the samples from the same class. To evaluate its efficacy in mitigating the memorization of pseudo-label noise, we conducted a comparative experiment. The results, depicted in Fig.~\ref{fig:CL_acc}, illustrate the accuracy of the selected pseudo labels. It is evident that \clshortname{} significantly enhances the precision of the selection process, highlighting its effectiveness in enhancing the overall performance of \selectshortname{}.

\mytext{Effect of Hyper-parameters.}
We conduct an evaluation to assess the sensitivity of three key hyper-parameters in our method, \shortname, on both the Office and the VisDA-C datasets. These hyper-parameters are as follows:
(1) The confident sample selection ratio, denoted as $\gamma$,
(2) The temperature parameter, $\tau$, in the $\mathcal{L}_{CL}$,
(3) The number of nearest neighbors, denoted as $K$.
To ensure a fair comparison, all other parameters were kept at their default values. 

The results, depicted in Fig.~\ref{fig:param}, indicate that these hyper-parameters have low sensitivity to the model's performance. This characteristic is highly advantageous in practical applications. Although the choice of $\gamma$ on the VisDA-C dataset is more critical, as it depends on both the performance of the source model and the characteristics of the target dataset, our approach consistently outperforms prior methods across a wide range of $\gamma$ values, ranging from 0.5 to 0.9.

It is important to note that the default training parameters we employ may not be optimal for the VisDA-C dataset. Nevertheless, \shortname{} demonstrates relative stability when confronted with variations in these parameters. This allows us to select a broad range of parameters for testing purposes while maintaining consistent training parameters across multiple datasets as much as possible.

\subsection{Visualization and Analysis}
In this section, we present visualizations of the experimental results, including the feature distribution and accuracy curves.

\mytext{Model Behavior Visualization.}
To assess the visual effectiveness of our proposed method, \shortname{}, we conduct a comprehensive analysis on the Ar$\to$Pr benchmark of the Office-Home dataset using t-SNE visualization~\cite{t-sne}. The results from two perspectives are compared: domain shift (Fig.~\ref{fig:tsne-a} vs. Fig.~\ref{fig:tsne-b}) and class-discriminative ability (Fig.~\ref{fig:tsne-c} vs. Fig.~\ref{fig:tsne-d}).

Fig.~\ref{fig:tsne-a} shows that the initial predictions on the target domain data exhibit clustering characteristics, but with significant class overlap. However, with the integration of \shortname{}, domain shifts are remarkably reduced, resulting in a more pronounced clustering structure (Fig.~\ref{fig:tsne-b}). Additionally, \shortname{} enables clear class boundaries, as evident in the visual comparison (Fig.~\ref{fig:tsne-c} vs. Fig.~\ref{fig:tsne-d}).

These visual insights provide compelling evidence for the discriminative ability of our proposed method, highlighting its effectiveness in enhancing domain adaptation performance.

\mytext{Training Curves.}
The accuracy progression of both \shortname{} (w/o \clshortname{}) and \shortname{} methods across two adaptation tasks is depicted in Fig.~\ref{fig:ap_acc} and Fig.~\ref{fig:vis_acc}. The results unequivocally demonstrate the rapid and consistent growth in accuracy achieved by our method. It is noteworthy that our approach exhibits swift convergence, typically within approximately 10 epochs, on both tasks, further substantiating its effectiveness and validating its performance.

\section{Conclusion}

In this study, we present a straightforward yet effective approach called \shortname{} SFUDA task. Our proposed \selectshortname{} module serves to filter out noisy pseudo labels, allowing for the utilization of high-quality pseudo labels that greatly enhance the adaptation process until convergence. Furthermore, the inclusion of the \clshortname{} ensures robust representation learning by enforcing class-wise contrastive consistency, thereby preventing the memorization of label noise. We conducted comprehensive experiments on three widely used SFUDA benchmarks, which demonstrate the superior performance of our approach in effectively denoising pseudo labels for SFUDA problems.

\mytext{Acknowledgement.}
This work is supported by the National Key R$\&$D Program of China (No. 2021ZD0110900).

{\small
\bibliographystyle{ieee_fullname}
\bibliography{main}
}

\end{document}


\title{Supplementary Material for CVPR 2023 paper \#1679}
\maketitle

\section{Further Implementation Details}

\subsection{Source model training}
We follow the standard protocol for source model training
\cite{SHOT}.
Specifically, we pretrain the source model $\phi_s$ 
by the standard cross-entropy loss with {\em label smoothing}:
\cite{LabelSmooth}: 
\begin{align}
    \mathcal{L}_{ce}=-\mathop{\mathbb{E}}_{(x_s,y_s)\in \mathcal{X}_s\times\mathcal{Y}_s}\sum_{c=1}^C T_c^{ls}\log\delta_c(\phi_s(x_s)), \\
    T_c^{ls}=(1-\alpha)T_c+\frac{\alpha}{C},
\end{align}
where 
$T_c$ is the $c$-th element of one-hot label vector,
$C$ is the total number of categories,
and the smoothing parameter $\alpha = 0.1$. 

\subsection{Illustrations of Ablation study}
\myparagraph{Quality of pseudo-label selection.}
We give a detailed description of the uncertainty estimation of pseudo-labels that we compare in the ablation study section.

Given the pseudo-label $y_i$ of target sample $x_i$:

\begin{equation}
    Prob(y_i) = \phi_t(x_i)_{y_i}
\end{equation}

\begin{equation}
    Ent(y_i) = -\frac{\sum_{c=1}^C \delta(\phi_t(x_i))\log(\delta(\phi_t(x_i)))}{C}
\end{equation}

\begin{equation}
    Cossim(y_i) = 1-\frac{g_t(x_i)\cdot\mu^{(1)}_{y_i}}{\Vert g_t(x_i)\Vert \Vert \mu^{(1)}_{y_i} \Vert}
\end{equation}

As for the comparison with CoWA-JMDS \cite{CoWA-JMDS}, JMDS score is their sample-wise confidence.

We select top-$\gamma \%$ ratio (80\% in our experiment on VisDA-C) of confident samples for each class every epoch and report their pseudo-label accuracy.




{\small
\bibliographystyle{ieee_fullname}
\bibliography{egbib}
}

%% file: tables/sota_visda.tex
\begin{table*}[ht]
\centering
\caption{Classification accuracy (\%) on {\bf VisDA-C} benchmark. S$\to$R denotes Synthesis$\to$Real task. Backbone: ResNet-101. \XSolidBrush: UDA methods; 
\Checkmark: SFUDA methods. Best results under SFUDA setting are shown in \textbf{bold} font. 
}
\scalebox{0.8}{
\begin{tabular}{c|c|cccccccccccc|c}
\toprule
Method (S$\to$R)&SF&plane&bcycl&bus&car&horse&knife&mcycl&person&plant&sktbrd&train&truck&Avg.\\
\midrule
\xchen{CDAN~\cite{CDAN}}&\XSolidBrush&85.2& 66.9& 83.0& 50.8& 84.2& 74.9& 88.1& 74.5& 83.4& 76.0& 81.9& 38.0& 73.9\\
\xchen{MCC~\cite{MCC}}&\XSolidBrush&88.7 &80.3& 80.5& 71.5& 90.1& 93.2& 85.0& 71.6& 89.4& 73.8& 85.0& 36.9 &78.8\\
BCDM~\cite{BCDM}&\XSolidBrush&95.1&87.6&81.2&73.2&92.7&95.4&86.9&82.5&95.1&84.8&88.1&39.5&83.4\\
FixBi ~\cite{FIxBi}&\XSolidBrush&96.1&87.8&90.5&90.3&96.8&95.3&92.8&88.7&97.2&94.2&90.9&25.7&87.2\\
CAN ~\cite{CAN}&\XSolidBrush&97.0&87.2&82.5&74.3&97.8&96.2&90.8&80.7&96.6&96.3&87.5&59.9&87.2\\

\midrule
source model only &\xchen{-}&56.9& 22.6& 49.5& 75.8& 62.5& 6.7& 83.3& 27.2& 58.3& 42.0& 82.0& 6.4 &47.8\\
3C-GAN ~\cite{3C-GAN}&\Checkmark&94.8&73.4&68.8&74.8&93.1&95.4&88.6&84.7&89.1&84.7&83.5&48.1&81.6\\
U-SFAN~\cite{U-SFAN}&\Checkmark& 94.9& 87.4& 78.0& 56.4& 93.8& 95.1& 80.5& 79.9& 90.1& 90.1& 85.3& 60.4& 82.7\\
DIPE ~\cite{DIPE}&\Checkmark&95.2&87.6&78.8&55.9&93.9&95.0&84.1&81.7&92.1&88.9&85.4&58.0&83.1\\
HCL~\cite{HCL}&\Checkmark&93.3&85.4&80.7&68.5&91.0&88.1&86.0&78.6&86.6&88.8&80.0&74.7&83.5\\
VDM-DA\cite{VDM-DA}&\Checkmark& 96.9& 90.0& 80.0& 64.4& \textbf{96.8}& 96.4& 86.7& 83.3& 96.2& 87.9& 89.8& 54.7& 85.3\\
NRC ~\cite{NRC} &\Checkmark&96.8& \textbf{91.3}& 82.4& 62.4& 96.2& 95.9& 86.1& 80.6& 94.8& 94.1& 90.4& 59.7 &85.9\\
SSNLL~\cite{SSNLL}&\Checkmark&\textbf{98.3}& 88.2& 88.1& 79.7& 96.5& 90.4& \textbf{92.5}& 81.6& 92.9 &86.3& 91.9& 45.4 &86.0\\
\xchen{AdaContrast~\cite{AdaContrast}}&\Checkmark&97.0& 84.7& 84.0& 77.3& 96.7& 93.8& 91.9& 84.8& 94.3& 93.1& \textbf{94.1}& 49.7& 86.8\\
CoWA-JMDS~\cite{CoWA-JMDS}&\Checkmark&96.2&89.7&83.9&73.8&96.4&97.4&89.3&86.8&94.6&92.1&88.7&53.8&86.9 \\

\xchen{CDCL~\cite{CDCL}}&\Checkmark&97.3& 90.5& 83.2& 59.9& 96.4& \textbf{98.4}& 91.5& 85.6& 96.0& 95.8& 92.0& 63.8& 87.5\\

D-MCD ~\cite{D-MCD}&\Checkmark&97.0&88.0&\textbf{90.0}&\textbf{81.5}&95.6&98.0&86.2&\textbf{88.7}&94.6&92.7&83.7&53.1&87.5\\

\midrule
SHOT~\cite{SHOT}&\Checkmark&94.3&88.5&80.1&57.3&93.1&94.9&80.7&80.3&91.5&89.1&86.3&58.2&82.9\\
{\bf \shortname} &\Checkmark& 97.0 & 90.4& 82.6 & 65.0 & 96.7 & 96.7 & 91.0 & 87.0 & \textbf{96.8} & \textbf{96.5} & 89.2 & \textbf{75.0}& \textbf{88.7} \\

\bottomrule
\end{tabular}
}
\label{table:Visda-C}
\end{table*}

%% file: tables/sota_office-home.tex
\begin{table*}[ht]
\caption{Classification accuracy (\%) on {\bf Office-Home} benchmark. Backbone: ResNet-50. \XSolidBrush: UDA methods; \Checkmark: SFUDA methods. Best results under SFUDA setting are shown in \textbf{bold} font.
}
\centering
\scalebox{0.7}{
\begin{tabular}{c|c|cccccccccccc|c}
\toprule
Method & SF &Ar$\to$Cl& Ar$\to$Pr & Ar$\to$Rw & Cl$\to$Ar & Cl$\to$Pr & Cl$\to$Rw & Pr$\to$Ar & Pr$\to$Cl & Pr$\to$Rw & Rw$\to$Ar & Rw$\to$Cl & Rw$\to$Pr & Avg.\\
\midrule
\xchen{CDAN~\cite{CDAN}}&\XSolidBrush&50.7& 70.6& 76.0& 57.6& 70.0& 70.0& 57.4& 50.9& 77.3& 70.9& 56.7& 81.6& 65.8\\
\xchen{MDD~\cite{MDD}}&\XSolidBrush&54.9&73.7&77.8&60.0&71.4&71.8&61.2&53.6&78.1&72.5&60.2&82.3&68.1\\
RSDA~\cite{RSDA}&\XSolidBrush&53.2&77.7&81.3&66.4&74.0&76.5&67.9&53.0&82.0&75.8&57.8&85.4&70.9\\
SRDC~\cite{SRDC}&\XSolidBrush&52.3&76.3&81.0&69.5&76.2&78.0&68.7&53.8&81.7&76.3&57.1&85.0&71.3\\
FixBi~\cite{FIxBi}&\XSolidBrush&58.1&77.3&80.4&67.7&79.5&78.1&65.8&57.9&81.7&76.4&62.9&86.7&72.7\\

\midrule
source model only&\xchen{-}& 44.0&65.9&74.0&52.1&61.0&65.4&52.2&40.9&72.9&64.1&45.3&77.9&59.6\\
VDM-DA~\cite{VDM-DA}&\Checkmark&59.3& 75.3& 78.3& 67.6& 76.0& 75.9& 68.8& 57.7& 79.6& 74.0& 61.1& 83.6& 71.4\\
U-SFAN~\cite{U-SFAN}&\Checkmark& 57.8& 77.8& 81.6& 67.9& 77.3& 79.2& 67.2& 54.7& 81.2& 73.3& 60.3& 83.9& 71.9\\
NRC ~\cite{NRC}&\Checkmark& 57.7& \textbf{80.3}& 82.0& 68.1& 79.8& 78.6& 65.3& 56.4& 83.0& 71.0& 58.6&\textbf{85.6} & 72.2\\
D-MCD ~\cite{D-MCD}&\Checkmark&\textbf{59.4}&78.9&80.2&67.2&79.3&78.6&65.3&55.6&82.2&73.3&\textbf{62.8}&83.9&72.2\\
CoWA-JMDS~\cite{CoWA-JMDS}&\Checkmark&56.9&78.4&81.0&69.1&80.0&79.9&67.7&\textbf{57.2}&82.4&72.8&60.5&84.5&72.5\\
DIPE~\cite{DIPE}&\Checkmark&56.5&79.2&80.7&\textbf{70.1}&79.8&78.8&67.9&55.1&\textbf{83.5}&74.1&59.3&84.8 &72.5\\
\midrule
SHOT~\cite{SHOT}&\Checkmark&57.1&78.1&81.5&68.0&78.2&78.1&67.4&54.9&82.2&73.3&58.8&84.3&71.8\\

{\bf \shortname}& \Checkmark&57.3&79.5& \textbf{82.4}& 68.2& \textbf{81.0}&\textbf{80.2}&\textbf{69.3}& 55.4& 82.3& \textbf{75.3}&59.2&85.5& \textbf{73.0}\\

\bottomrule
\end{tabular}
}
\label{table:office-home}
\end{table*}

%% file: tables/sota_office.tex
\begin{table}[ht]
\caption{Classification accuracy (\%) on {\bf Office} benchmark. Backbone: ResNet-50. \XSolidBrush: UDA methods; 
\Checkmark: SFUDA methods. The best results under the SFUDA setting are shown in \textbf{bold} font.
}
\centering
\scalebox{0.6}{
\begin{tabular}{c|c|cccccc|c}
\toprule
    Method & SF & A$\to$D& A$\to$W & D$\to$A & D$\to$W & W$\to$A & W$\to$D & Avg.  \\
\midrule
\xchen{CDAN~\cite{CDAN}}&\XSolidBrush&92.9&94.1&71.0&98.6&69.3&100.0&87.7\\
\xchen{MDD~\cite{MDD}}&\XSolidBrush&93.5&94.5&74.6&98.4&72.2&100.0&88.9\\
BCDM~\cite{BCDM}&\XSolidBrush&93.8&95.4&73.1&98.6&71.6&100.0&89.0\\
CAN~\cite{CAN}&\XSolidBrush&95.0&94.5&78.0&99.1&77.0&99.8&90.6\\
SRDC~\cite{SRDC}&\XSolidBrush&95.7&99.2&100.0&95.8&76.7&77.1&90.8\\
RSDA~\cite{RSDA}&\XSolidBrush&96.1&78.9&95.8&77.4&99.3&100.0&91.1\\

\midrule
source model only&\xchen{-}&82.5&76.9&61.9&95.2&62.7&98.0&79.5\\
SSNLL~\cite{SSNLL}&\Checkmark&88.4& 90.8& 75.9& 96.7& 76.1& 99.8& 88.0\\
U-SFAN~\cite{U-SFAN}&\Checkmark& 94.2& 92.8& 74.6& 98.0& 74.4& 99.0& 88.8\\
\xchen{CDCL\cite{CDCL}}&\Checkmark&94.4&92.1&\textbf{76.4}&98.5&74.1&\textbf{100.0}&89.3\\
NRC~\cite{NRC}&\Checkmark& 96.0& 90.8& 75.3& \textbf{99.0}& 75.0& 100.0& 89.4\\
3C-GAN~\cite{3C-GAN}&\Checkmark&92.7& 93.7& 75.3& 98.5&\textbf{77.8}&99.8&89.6\\
VDM-DA~\cite{VDM-DA}&\Checkmark&94.1& 93.2& 75.8& 98.0& 77.1& \textbf{100.0}& 89.7\\
HCL~\cite{HCL}&\Checkmark&94.7&92.5&75.9&98.2&77.7&\textbf{100.0}&89.8\\
D-MCD~\cite{D-MCD}&\Checkmark&94.1&93.5&\textbf{76.4}&98.8&76.4&\textbf{100.0}&89.9\\
DIPE~\cite{DIPE}& \Checkmark& \textbf{96.6}& 93.1& 75.5& 98.4& 77.2& 99.6& 90.1\\
CoWA-JMDS~\cite{CoWA-JMDS}&\Checkmark &94.4& \textbf{95.2}& 76.2& 98.5& 77.6& 99.8& \textbf{90.3}\\

\midrule
SHOT~\cite{SHOT}&\Checkmark& 94.0& 90.1& 74.7& 98.4& 74.3& 99.9& 88.6\\
{\bf \shortname}&\Checkmark& 96.2& 93.7& 75.0& 98.4& 76.6& 99.8& 89.9\\

\bottomrule
\end{tabular}
}
\label{table:office}
\end{table}

%% file: tables/sota_domainnet.tex
\begin{table*}[!htbp]
\caption{\xchen{Classification accuracy (\%) on 7 domain shifts of {\bf DomainNet-126}. Backbone: ResNet-50. \XSolidBrush: UDA methods; \Checkmark: SFUDA methods. Best results under SFUDA setting are shown in \textbf{bold} font.
}}
\centering
\scalebox{0.7}{
\begin{tabular}{c|c|ccccccc|c}
\toprule
Method & SF &R$\to$C& R$\to$P & P$\to$C & Cl$\to$S & S$\to$P & R$\to$S & P$\to$R &Avg.\\
\midrule
MCC~\cite{MCC}&\XSolidBrush& 44.8& 65.7& 41.9& 34.9& 47.3& 35.3& 72.4& 48.9\\
\midrule
source model only&-& 54.8 & 62.4& 52.5 & 46.6& 50.6& 45.9& 75.1& 55.4\\
Tent~\cite{TENT}&\Checkmark&58.5& 65.7& 57.9& 48.5& 52.4& 54.0& 67.0& 57.7\\
AdaContrast~\cite{AdaContrast}&\Checkmark&\textbf{70.2}& \textbf{69.8}& \textbf{68.6}&58.0& 65.9&\textbf{61.5}&80.5&67.8\\

\midrule
SHOT~\cite{SHOT}&\Checkmark&67.7 &68.4& 66.9&60.1& 66.1& 59.9& 80.8& 67.1\\
{\bf \shortname}& \Checkmark&68.6& 69.5& 67.6& \textbf{60.9}& \textbf{66.8}&\textbf{61.5}& \textbf{80.9}& \textbf{68.0}\\

\bottomrule
\end{tabular}
}
\label{table:domainnet}
\end{table*}

%% file: tables/ablation_loss.tex
\begin{table}[ht]
\centering
\caption{Ablation results (\%) of sub-components on the three datasets.}
\label{table:ablation_loss}
\scalebox{0.7}{
\begin{tabular}{c|cc|ccc }
\toprule
Methods & \selectshortname & \clshortname & Office & Office-Home & VisDA-C\\
\midrule
1 & & & 88.6 & 71.8 & 82.9\\
2 & \Checkmark & & 89.5 & 72.7 & 84.2 \\
3 & \Checkmark & \Checkmark & \textbf{89.9} & \textbf{73.0} & \textbf{88.7} \\
\bottomrule
\end{tabular}
}
\end{table}

%% file: tables/ablation_pseudo.tex
\begin{table}[ht]
\centering
\caption{Comparisions of accuracy (\%) on Ar$\to$Pr task in Office-Home dataset and  S$\to$R task in VisDA-C dataset with various uncertainty estimation methods for pseudo-label selection. The best results are shown in \textbf{bold} font.}
\label{table:PL}
\scalebox{0.8}{
\begin{tabular}{cccccc}
\toprule
Task&  Ent & Prob & Cossim & APS(Ours) \\
\midrule
    Ar$\to$Pr&78.0 & 78.4& 78.5 & \textbf{79.5}\\
    S$\to$R&86.2 &86.0 &84.9 & \textbf{88.7}\\
\bottomrule
\end{tabular}
}
\end{table}

%% file: tables/ablation_pe.tex
\begin{table}[ht]
\centering
\caption{Impact of iterative refinement in the \selectshortname{} module on VisDA-C and Office Datasets.}
\label{tab:selection_times}
\scalebox{0.8}{
\begin{tabular}{ccc}
\toprule
Iteration & VisDA-C (\%) & Office (\%) \\
\midrule
    1 & 87.2 & 89.5 \\ 
    2(Ours)& \textbf{88.7} &\textbf{89.9} \\
    3 & 88.7 & 89.8 \\
\bottomrule
\end{tabular}
}
\end{table}

%% file: main.bbl
\begin{thebibliography}{10}\itemsep=-1pt

\bibitem{MADA}
Zhangjie Cao, Lijia Ma, Mingsheng Long, and Jianmin Wang.
\newblock {Multi-Adversarial Domain Adaptation}.
\newblock In {\em AAAI}, pages 3934--3941, 2018.

\bibitem{AdaContrast}
Dian Chen, Dequan Wang, and Trevor Darrell.
\newblock {Contrastive Test-Time Adaptation}.
\newblock In {\em CVPR}, 2022.

\bibitem{SSNLL}
Weijie Chen, Luojun Lin, Shicai Yang, Di Xie, Shiliang Pu, Yueting Zhuang, and
  Wenqi Ren.
\newblock {Self-Supervised Noisy Label Learning for Source-Free Unsupervised
  Domain Adaptation}.
\newblock In {\em IROS}, 2022.

\bibitem{MOCOV2}
Xinlei Chen, Haoqi Fan, Ross Girshick, and Kaiming He.
\newblock {Improved Baselines with Momentum Contrastive Learning}, 2020.

\bibitem{D-MCD}
Tong Chu, Yahao Liu, Jinhong Deng, Wen Li, and Lixin Duan.
\newblock {Denoised Maximum Classifier Discrepancy for Source-Free Unsupervised
  Domain Adaptation}.
\newblock In {\em AAAI}, pages 472--480, 2022.

\bibitem{DANN}
Yaroslav Ganin, Evgeniya Ustinova, Hana Ajakan, Pascal Germain, Hugo
  Larochelle, Fran{\c{c}}ois Laviolette, Mario Marchand, and Victor Lempitsky.
\newblock {Domain-adversarial training of neural networks}.
\newblock {\em JMLR}, 17:189--209, 2017.

\bibitem{RL-NL}
Aritra Ghosh, Himanshu Kumar, and P.~S. Sastry.
\newblock {Robust loss functions under label noise for deep neural networks}.
\newblock In {\em AAAI}, pages 1919--1925, 2017.

\bibitem{MMD}
Arthur Gretton, Karsten~M. Borgwardt, Malte~J. Rasch, Bernhard Sch¨olkopf, and
  Alexander Smola{\S}.
\newblock {A Kernel Two-Sample Test}.
\newblock {\em Journal of Machine Learning Research}, 13:723--773, 2012.

\bibitem{RSDA}
Xiang Gu, Jian Sun, and Zongben Xu.
\newblock {Spherical Space Domain Adaptation with Robust Pseudo-Label Loss}.
\newblock In {\em CVPR}, pages 9098--9107, 2020.

\bibitem{Resnet}
Kaiming He, Xiangyu Zhang, Shaoqing Ren, and Jian Sun.
\newblock {Deep residual learning for image recognition}.
\newblock In {\em CVPR}, pages 770--778, 2016.

\bibitem{GAM}
Haoshuo Huang, Qixing Huang, and Philipp Kr{\"{a}}henb{\"{u}}hl.
\newblock {Domain Transfer Through Deep Activation Matching}.
\newblock In {\em ECCV}, 2018.

\bibitem{HCL}
Jiaxing Huang, Dayan Guan, Aoran Xiao, and Shijian Lu.
\newblock {Model Adaptation: Historical Contrastive Learning for Unsupervised
  Domain Adaptation without Source Data}.
\newblock In {\em NIPS}, pages 1--15, 2021.

\bibitem{MCC}
Ying Jin, Ximei Wang, Mingsheng Long, and Jianmin Wang.
\newblock {Minimum Class Confusion for Versatile Domain Adaptation}.
\newblock In {\em ECCV}, 2020.

\bibitem{CAN}
Guoliang Kang, Lu Jiang, Yi Yang, and Alexander~G. Hauptmann.
\newblock {Contrastive adaptation network for unsupervised domain adaptation}.
\newblock In {\em CVPR}, pages 4888--4897, 2019.

\bibitem{SupCon}
Prannay Khosla, Piotr Teterwak, Chen Wang, Aaron Sarna, Yonglong Tian, Phillip
  Isola, Aaron Maschinot, Ce Liu, and Dilip Krishnan.
\newblock {Supervised contrastive learning}.
\newblock In {\em NIPS}, pages 1--23, 2020.

\bibitem{CoWA-JMDS}
Jonghyun Lee, Dahuin Jung, Junho Yim, and Sungroh Yoon.
\newblock {Confidence Score for Source-Free Unsupervised Domain Adaptation}.
\newblock In {\em ICML}, pages 12365--12377, 2022.

\bibitem{MUDA}
Joon~Ho Lee and Gyemin Lee.
\newblock {Unsupervised domain adaptation based on the predictive uncertainty
  of models}.
\newblock {\em Neurocomputing}, 520:183--193, 2023.

\bibitem{DivideMix}
Junnan Li, Richard Socher, and Steven C.~H. Hoi.
\newblock {DivideMix: Learning with Noisy Labels as Semi-supervised Learning}.
\newblock In {\em ICLR}, pages 1--14, 2020.

\bibitem{3C-GAN}
Rui Li, Qianfen Jiao, Wenming Cao, Hau~San Wong, and Si Wu.
\newblock {Model Adaptation: Unsupervised Domain Adaptation without Source
  Data}.
\newblock In {\em CVPR}, pages 9638--9647, 2020.

\bibitem{BCDM}
Shuang Li, Fangrui Lv, Binhui Xie, Chi~Harold Liu, Jian Liang, and Chen Qin.
\newblock {Bi-Classifier Determinacy Maximization for Unsupervised Domain
  Adaptation}.
\newblock In {\em AAAI}, pages 8455--8464, 2021.

\bibitem{Sel-CL}
Shikun Li, Xiaobo Xia, Shiming Ge, and Tongliang Liu.
\newblock {Selective-Supervised Contrastive Learning with Noisy Labels}.
\newblock In {\em CVPR}, pages 316--325, 2022.

\bibitem{SHOT}
Jian Liang, Dapeng Hu, and Jiashi Feng.
\newblock {Do we really need to access the source data? source hypothesis
  transfer for unsupervised domain adaptation}.
\newblock In {\em ICML}, pages 5984--5995, 2020.

\bibitem{CDAN}
Mingsheng Long, Zhangjie Cao, Jianmin Wang, and Michael~I. Jordan.
\newblock {Conditional adversarial domain adaptation}.
\newblock In {\em NIPS}, pages 1640--1650, 2018.

\bibitem{LabelSmooth}
Rafael M{\"{u}}ller, Simon Kornblith, and Geoffrey Hinton.
\newblock {When Does Label Smoothing Help ?}
\newblock In {\em Advances in neural information processing systems}, 2019.

\bibitem{FIxBi}
Jaemin Na, Heechul Jung, Hyung~Jin Chang, and Wonjun Hwang.
\newblock {FixBi: Bridging Domain Spaces for Unsupervised Domain Adaptation}.
\newblock In {\em CVPR}, pages 1094--1103, 2021.

\bibitem{MOIT}
Diego Ortego, Eric Arazo, Paul Albert, Noel~E. O'Connor, and Kevin McGuinness.
\newblock {Multi-Objective Interpolation Training for Robustness to Label
  Noise}.
\newblock In {\em CVPR}, pages 6602--6611, 2021.

\bibitem{LNLC}
Giorgio Patrini, Alessandro Rozza, Aditya~Krishna Menon, Richard Nock, and
  Lizhen Qu.
\newblock {Making deep neural networks robust to label noise: A loss correction
  approach}.
\newblock In {\em CVPR}, pages 2233--2241, 2017.

\bibitem{DomainNet}
Xingchao Peng, Qinxun Bai, Xide Xia, Zijun Huang, Kate Saenko, and Bo Wang.
\newblock {Moment matching for multi-source domain adaptation}.
\newblock In {\em ICCV}, 2019.

\bibitem{Visda-c}
Xingchao Peng, Ben Usman, Neela Kaushik, Judy Hoffman, Dequan Wang, and Kate
  Saenko.
\newblock {VisDA: The Visual Domain Adaptation Challenge}, 2017.

\bibitem{SimNet}
Pedro~O. Pinheiro.
\newblock {Unsupervised Domain Adaptation with Similarity Learning}.
\newblock In {\em CVPR}, pages 8004--8013, 2018.

\bibitem{CPGA}
Zhen Qiu, Yifan Zhang, Hongbin Lin, Shuaicheng Niu, Yanxia Liu, Qing Du, and
  Mingkui Tan.
\newblock {Source-free Domain Adaptation via Avatar Prototype Generation and
  Adaptation}.
\newblock In {\em IJCAI}, pages 2921--2927, 2021.

\bibitem{U-SFAN}
Subhankar Roy, Martin Trapp, Andrea Pilzer, Juho Kannala, Nicu Sebe, Elisa
  Ricci, and Arno Solin.
\newblock {Uncertainty-guided Source-free Domain Adaptation}.
\newblock In {\em ECCV}, 2022.

\bibitem{office-31}
Kate Saenko, Brian Kulis, Mario Fritz, and Trevor Darrell.
\newblock {Adapting visual category models to new domains.pdf}.
\newblock In {\em ECCV}, pages 213--226, 2012.

\bibitem{MME}
Kuniaki Saito, Donghyun Kim, Stan Sclaroff, Trevor Darrell, and Kate Saenko.
\newblock {Semi-supervised domain adaptation via minimax entropy}.
\newblock In {\em ICCV}, 2019.

\bibitem{CORAL}
Baochen Sun, Jiashi Feng, and Kate Saenko.
\newblock {Return of frustratingly easy domain adaptation}.
\newblock In {\em AAAI}, pages 2058--2065, 2016.

\bibitem{D-CORAL}
Baochen Sun and Kate Saenko.
\newblock {Deep CORAL: Correlation alignment for deep domain adaptation}.
\newblock In {\em ECCV Workshops}, pages 443--450, 2016.

\bibitem{SRDC}
Hui Tang, Ke Chen, and Kui Jia.
\newblock {Unsupervised domain adaptation via structurally regularized deep
  clustering}.
\newblock In {\em CVPR}, pages 8722--8732, 2020.

\bibitem{VDM-DA}
Jiayi Tian, Jing Zhang, Wen Li, and Dong Xu.
\newblock {VDM-DA: Virtual Domain Modeling for Source Data-free Domain
  Adaptation}.
\newblock {\em TCSVT}, 8215(c):3749--3760, 2021.

\bibitem{ASOS}
Yi~Hsuan Tsai, Wei~Chih Hung, Samuel Schulter, Kihyuk Sohn, Ming~Hsuan Yang,
  and Manmohan Chandraker.
\newblock {Learning to Adapt Structured Output Space for Semantic
  Segmentation}.
\newblock In {\em CVPR}, 2018.

\bibitem{t-sne}
Laurens van~der Maaten and Geoffrey Hinton.
\newblock {Visualizing Data using t-SNE Laurens}.
\newblock {\em Journal of Machine Learning Research}, 9(11):187--202, 2008.

\bibitem{office-home}
Hemanth Venkateswara, Jose Eusebio, Shayok Chakraborty, and Sethuraman
  Panchanathan.
\newblock {Deep hashing network for unsupervised domain adaptation}.
\newblock In {\em CVPR}, pages 5385--5394, 2017.

\bibitem{TENT}
Dequan Wang, Evan Shelhamer, Shaoteng Liu, Bruno Olshausen, and Trevor Darrell.
\newblock {TENT: FULLY TEST-TIME ADAPTATION BY ENTROPY MINIMIZATION}.
\newblock In {\em ICLR}, 2021.

\bibitem{DIPE}
Fan Wang, Zhongyi Han, Yongshun Gong, and Yilong Yin.
\newblock {Exploring Domain-Invariant Parameters for Source Free Domain
  Adaptation}.
\newblock In {\em CVPR}, pages 7151--7160, 2022.

\bibitem{CDCL}
Rui Wang, Zuxuan Wu, Zejia Weng, Jingjing Chen, Guo~Jun Qi, and Yu~Gang Jiang.
\newblock {Cross-domain Contrastive Learning for Unsupervised Domain
  Adaptation}.
\newblock {\em TMM}, pages 1--11, 2022.

\bibitem{A2Net}
Haifeng Xia, Handong Zhao, and Zhengming Ding.
\newblock {Adaptive Adversarial Network for Source-free Domain Adaptation}.
\newblock In {\em ICCV}, pages 9010--9019, 2021.

\bibitem{NRC}
Shiqi Yang, Yaxing Wang, Joost van~de Weijer, Luis Herranz, and Shangling Jui.
\newblock {Exploiting the Intrinsic Neighborhood Structure for Source-free
  Domain Adaptation}.
\newblock In {\em Advances in neural information processing systems}, pages
  1--13, 2021.

\bibitem{mixup}
Hongyi Zhang, Moustapha Cisse, Yann~N. Dauphin, and David Lopez-Paz.
\newblock {MixUp: Beyond empirical risk minimization}.
\newblock In {\em ICLR}, pages 1--13, 2018.

\bibitem{MDD}
Yuchen Zhang, Tianle Liu, Mingsheng Long, and Michael~I Jordan.
\newblock {Bridging Theory and Algorithm for Domain Adaptation}.
\newblock In {\em ICML}, pages 7404--7413., 2019.

\bibitem{MADAN}
Sicheng Zhao, Bo Li, Pengfei Xu, Xiangyu Yue, Guiguang Ding, and Kurt Keutzer.
\newblock {MADAN: Multi-source Adversarial Domain Aggregation Network for
  Domain Adaptation}.
\newblock {\em International Journal of Computer Vision}, 129:2399--2424, 2021.

\bibitem{SACR}
Xiaodi Zhu, Yanfeng Li, Jia Sun, Houjin Chen, and Jinlei Zhu.
\newblock {Learning with noisy labels method for unsupervised domain adaptive
  person re-identification}.
\newblock {\em Neurocomputing}, 452:78--88, 2021.

\end{thebibliography}
